\documentclass[review,12pt]{elsarticle}

\usepackage{amssymb}
\usepackage{amsthm}
\usepackage{enumitem}
\usepackage{amsmath}
\usepackage{subcaption}
\usepackage{hyperref}
\usepackage{multirow}

\usepackage{graphicx}

\usepackage[figuresright]{rotating}

\begin{document}

\begin{frontmatter}

\makeatletter
\def\ps@pprintTitle{%
 \let\@oddhead\@empty
 \let\@evenhead\@empty
 \let\@oddfoot\@empty
 \let\@evenfoot\@empty}
\makeatother

\title{Inductive Conformal Prediction under Data Scarcity: \\ Exploring the Impacts of Nonconformity Measures}


\author[label1]{\texorpdfstring{Yuko Kato}{Yuko Kato}}
\ead{Y.Kato@tudelft.nl}
\author[label1]{David M.J. Tax}
\ead{D.M.J.Tax@tudelft.nl}
\author[label2]{Marco Loog}
\ead{Marco.Loog@ru.nl}

\affiliation[label1]{organization = {Pattern Recognition and Bio-informatics Group, Delft University of Technology}, addressline={Van Mourik Broekmanweg 6},city = {Delft}, postcode={2628 XE}, country = {The Netherlands}}
\affiliation[label2]{organization = {Institute for Computing and Information Sciences, Radboud University}, addressline={Toernooiveld 212, 6525~EC}, city = {Nijmegen}, country = {The Netherlands}}

% \cortext[cor1]{Corresponding author}

\begin{abstract}
Conformal prediction, which makes no distributional assumptions about the data, has emerged as a powerful and reliable approach to uncertainty quantification in practical applications.  The nonconformity measure used in conformal prediction quantifies how a test sample differs from the training data and the effectiveness of a conformal prediction interval may depend heavily on the precise measure employed. The impact of this choice has, however, not been widely explored, especially when dealing with limited amounts of data. The primary objective of this study is to evaluate the performance of various nonconformity measures (absolute error-based, normalized absolute error-based, and quantile-based measures) in terms of validity and efficiency when used in inductive conformal prediction.  The focus is on small datasets, which is still a common setting in many real-world applications. Using synthetic and real-world data, we assess how different characteristics---such as dataset size, noise, and dimensionality---can affect the efficiency of conformal prediction intervals. Our results show that although there are differences, no single nonconformity measure consistently outperforms the others, as the effectiveness of each nonconformity measure is heavily influenced by the specific nature of the data. Additionally, we found that increasing dataset size does not always improve efficiency, suggesting the importance of fine-tuning models and, again, the need to carefully select the nonconformity measure for different applications.

\end{abstract}


\end{frontmatter}

% \linenumbers


\section{Introduction}
\label{Introduction}
With the increasing popularity of machine learning (ML) models, concerns about the accuracy of their predictions have sparked renewed interest in uncertainty quantification (UQ). Conformal prediction (CP) has been proposed as a powerful and reliable UQ approach, benefiting from its distribution-free assumption on input data. This assumption is particularly useful in practical applications where the data distribution is unknown, as it helps mitigate the risk of model misspecification, which can negatively impact the quality of uncertainty estimates \cite{Xu2021-wj,Fridovich-Keil2020-lh,Kato2023-ns}. Various researchers have demonstrated the high efficiency of CP methods while confirming their guaranteed validity across multiple domains \cite{Sun2023-ra,Olsson2022-fy}.

A key component of CP is the selection of the Nonconformity Measure (NCM), which quantifies how different a test sample is from the training data \cite{Shafer2007-ca}. The effectiveness of the conformal prediction intervals (PIs) can be highly dependent on the appropriateness of the selected NCM \cite{Papadopoulos2008-xu}.  Unfortunately, the impact of this choice is not well understood, particularly when dealing with limited amounts of data, e.g., sample sizes of a thousand or less, and multi-dimensional data where larger datasets are typically required \cite{Bansal2022-ng}. Generally, the existing literature offers little guidance on how different NCMs perform under such conditions.  This leaves a significant gap in our understanding of CP's effectiveness in real-world applications where data scarcity is common.

Our work investigates the impact of various data characteristics---such as dataset size, noise, and dimensionality---on CP performance in regression settings using Inductive Conformal Prediction (ICP) \cite{Papadopoulos2008-xu, Papadopoulos2002-bi}.  ICP is one of the two major categories of conformal predictors; the other being Transductive CP (TCP) \cite{Saunders1999-ts}. Specifically, we will simulate conditions with significantly smaller datasets compared to other studies.  This situation corresponds to many settings encountered in real-world applications. For comparison, we refer to the work by Linusson et al. \cite{Linusson2014-db}, who examined the effectiveness of CP on small datasets ($<$ 500 samples) in classification settings. Although their focus was classification and only one type of NCM was considered, their insights into small-sample behavior are relevant to our investigation of regression tasks, as it allows us to compare the impact of small datasets in both classification and regression.  

Through a series of experiments, we investigate the strengths and limitations of different NCMs under constrained conditions, with a focus on how data size affects CP performance in terms of validity and efficiency. Based on our findings, we aim to provide guidelines for selecting appropriate NCMs when applying CP in scenarios where data scarcity is one of the key challenges.

\section{Materials and Methods}
To compare the performance of different NCMs under various conditions, we conduct a series of regression experiments on synthetic and real-world data using ICP. We first assess the NCMs with synthetically generated data to understand their behavior under controlled conditions. Subsequently, we evaluate the NCMs on real-world data to examine their practical applicability and robustness. The experimental setup is outlined in \ref{subsec:experimental_setup} at the very end of this section.  The first six sections go through the data used, a description of ICP, the validity and efficiency as performance measures, the regression models considered, and the NCMs used.

\subsection{Data}
Synthetic data were generated capturing different complexities and noise structures. Let $X$ represent the input feature matrix, where $X \in \mathbb{R}^{n \times d}$ consists of $n$ samples and $d$ features, i.e., the data dimensionality is $d$. For the one-dimensional case ($d=1$), $X$ was uniformly distributed on the interval $[0,10]$. For cases with higher dimensionality ($d = 2, 5, 8$), $X$ was sampled from a standard normal distribution. The target variable $y$ was generated using a nonlinear transformation of the input data as follows:
\begin{equation*}
    y_{i} = \sum_{j=1}^{d} x_{i,j} \sin(x_{i,j}).
    \label{eq:placeholder}
\end{equation*}
% The functional form is used in various other studies \cite{Skafte2019-ib}. 

\begin{figure*}[htbp]
  \centering
  \begin{subfigure}{0.45\linewidth}
    \centering
    \includegraphics[width=\linewidth]{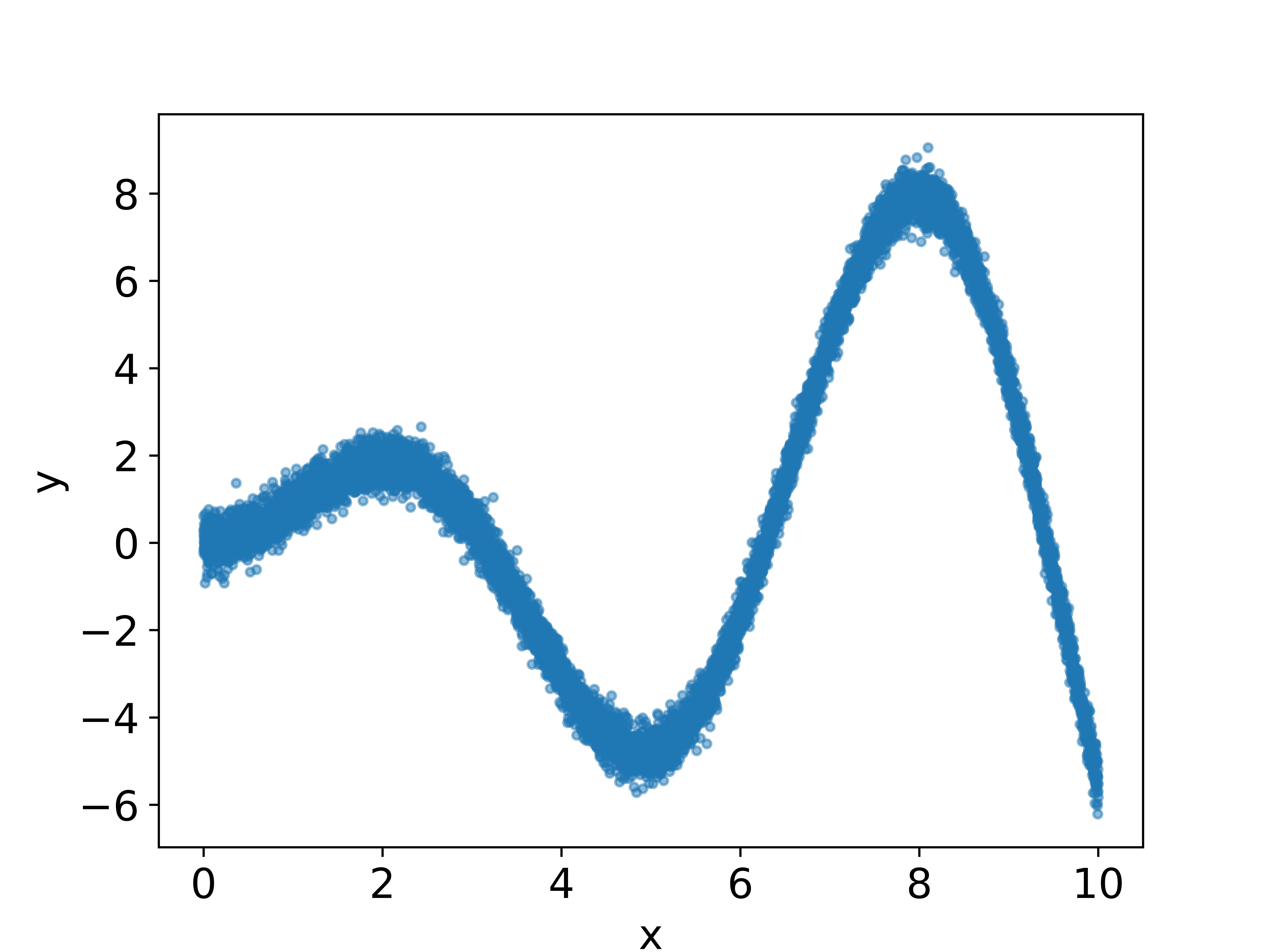}
    \caption{Homoscedastic Gaussian noise}
    \label{fig:Homoscedastic}
  \end{subfigure}
  % \hspace*{\fill}
  \begin{subfigure}{0.45\linewidth}
    \centering
    \includegraphics[width=\linewidth]{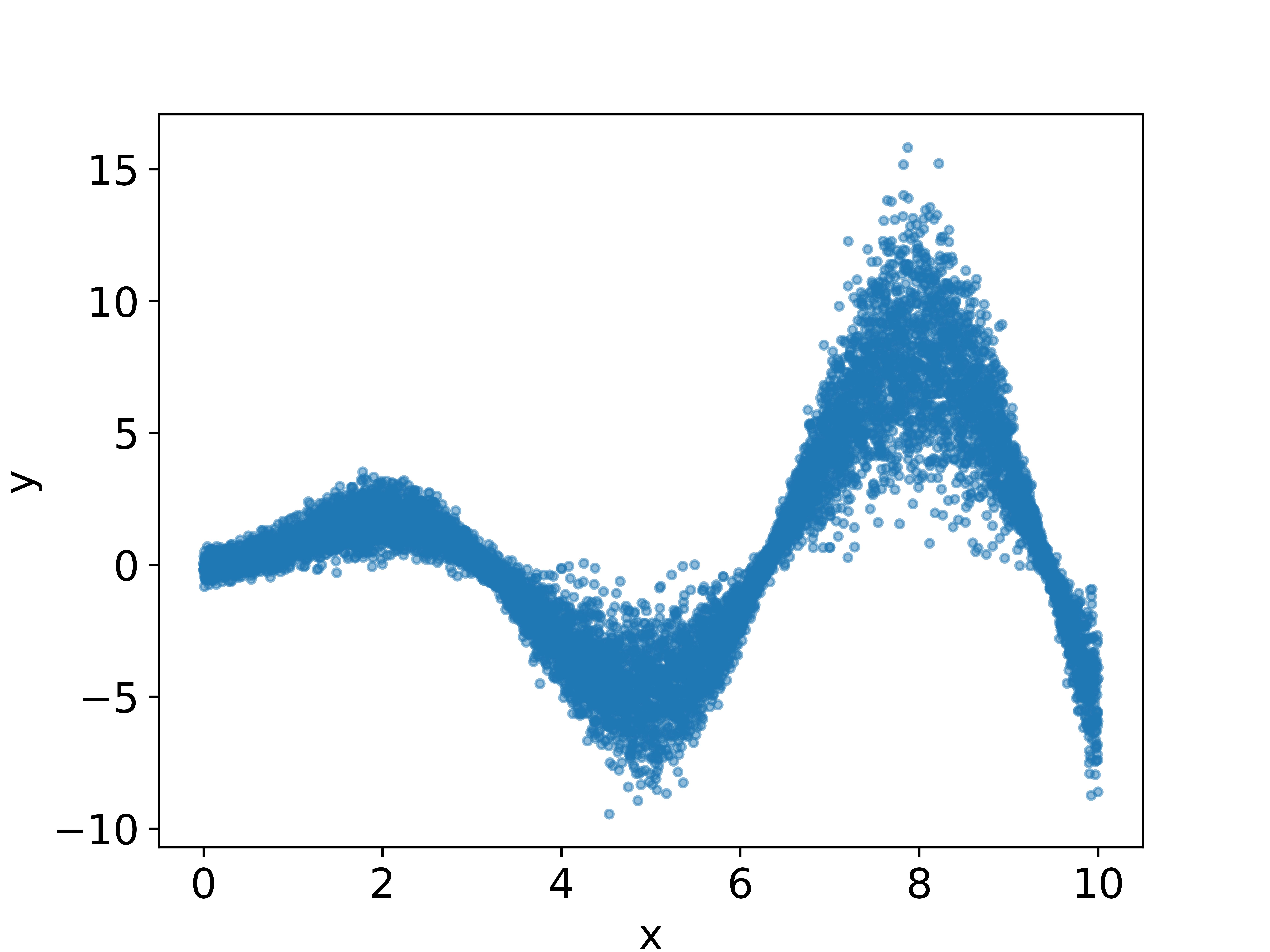}
    \caption{Heteroscedastic Gausian noise}
    \label{fig:Heteroscedastic}
  \end{subfigure}
  \vspace{0.5cm}
  \begin{subfigure}{0.45\linewidth}
    \centering
    \includegraphics[width=\linewidth]{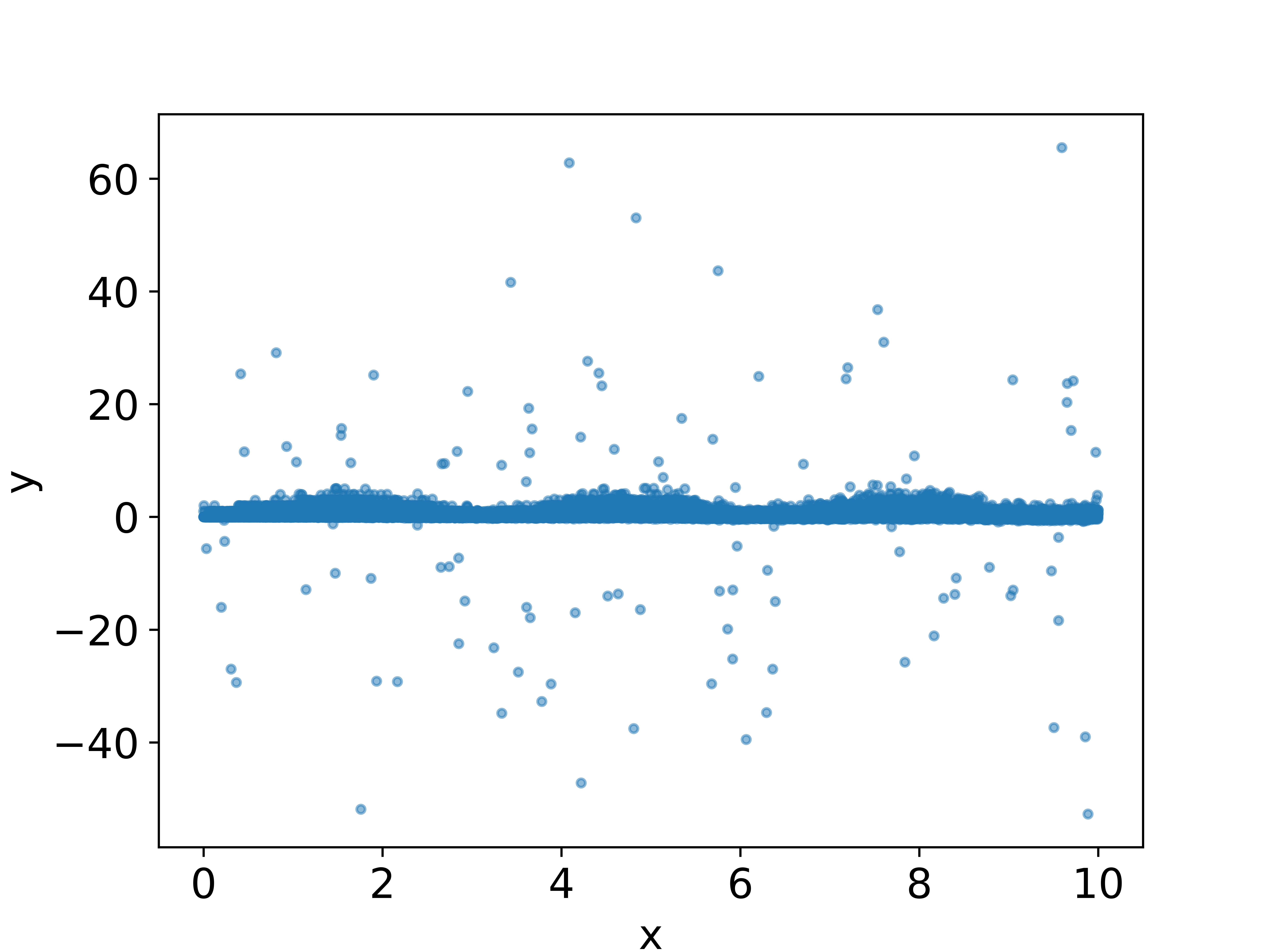}
    \caption{Heteroscedastic Non-Gaussian noise}
    \label{fig:Heteroscedastic2}
  \end{subfigure}
  % \hspace*{\fill}
  \begin{subfigure}{0.45\linewidth}
    \centering
    \includegraphics[width=\linewidth]{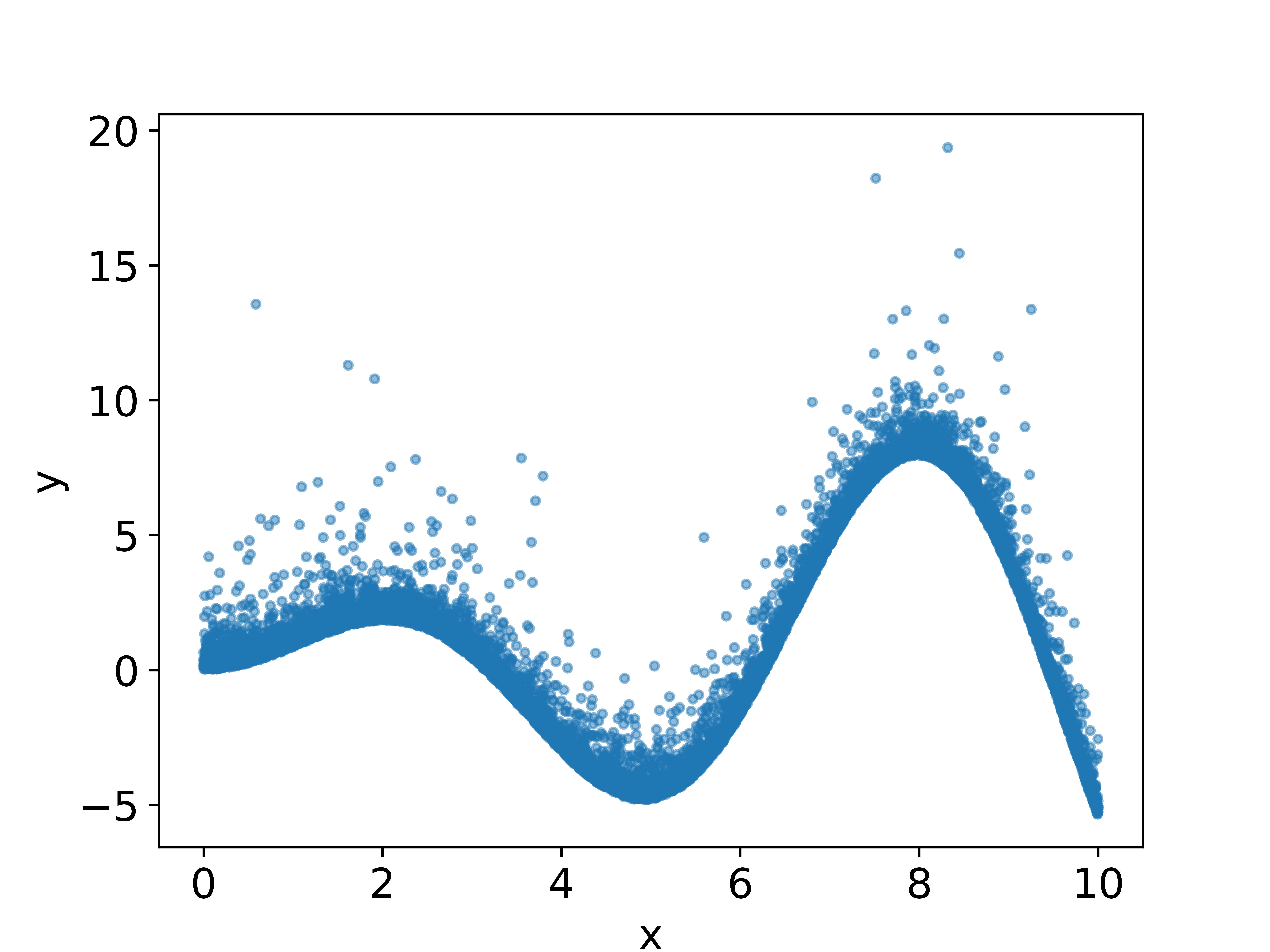}
    \caption{Right-skewed noise}
    \label{fig:Right}
  \end{subfigure}
  \caption{One dimensional synthetic data with different noise distribution (a) Homoscedastic Gaussian noise, (b) Heteroscedastic Gaussian noise, (c) Heteroscedastic non-Gaussian noise and (d) Homoscedastic right-skewed noise}
  \label{fig:data-combined}
\end{figure*}

In order to assess the impact of noise, we added different types of noise to the data. Specifically, we considered three noise types:
\begin{itemize}
    \item \textbf{Homoscedastic Gaussian noise}:\\  
    \( y_i =  \sum_{j=1}^{d} x_{i,j} \sin(x_{i,j}) + c \cdot \epsilon_{1i}  \), where \( \epsilon_{1i} \sim \mathcal{N}(0, 1) \),
    \item \textbf{Heteroscedastic Gaussian noise}:\\ \( y_i =  \sum_{j=1}^{d} x_{i,j} \sin(x_{i,j}) + c \cdot \epsilon_{1i} + c \cdot 
|y_i| \cdot \epsilon_{2i} \), where \( \epsilon_{1i}, \epsilon_{2i} \sim \mathcal{N}(0, 1) \),
    \item \textbf{Homoscedastic right-skewed noise}:\\ \( y_i =  \sum_{j=1}^{d} x_{i,j} \sin(x_{i,j}) + c \cdot \epsilon_{1i}  \), where \( \epsilon_{1i} \sim \text{LogNorm}(0, 1) \),
\end{itemize}

\noindent where LogNorm $(a,b)$ stands for the log-normal distribution with a mean of $a$ and a shape parameter $b$, and $c$ represents the fixed noise level.  Additionally, for the one-dimensional datasets ($d=1$), we included an additional dataset that was specifically used in the original work describing conformalized quantile regression\cite{Romano2019-ci}. This dataset was incorporated into our experiment to assess how NCMs perform on this specific dataset as well as to see the impact of small data size. The target variable $y$ for this specific dataset is sampled as:
\begin{itemize}
    \item \textbf{Heteroscedastic Non-Gaussian noise}:\\ \(y_i = \text{Pois}(\sin^2(x_i) + 0.1) + 0.03 x_i \epsilon_{1i} + 25 \mathbf{1}\{U_i < 0.01\} \epsilon_{2i}\), where \( \epsilon_{1i}, \epsilon_{2i} \sim \mathcal{N}(0, 1) \),
\end{itemize}
\noindent where Pois($\lambda$) is the Poisson distribution with mean $\lambda$. All one-dimensional data is illustrated in Figure \ref{fig:data-combined}.

To explore the impact of dataset size, synthetic datasets with varying sample sizes were generated, ranging from 100 to 1000 samples. This range allowed us to observe the models' behavior in smaller datasets, significantly smaller than the size of commonly used datasets in literature. 

Furthermore, real-world datasets (Concrete Compressive Strength ($n = 1.030, d = 9$), Airfoil Self-Noise ($n = 1.503, d = 5$), Combined Cycle Power Plant ($n = 9.568, d = 4$)) from the UCI repositories \cite{Kelly2023-nh} were used to evaluate each NCM's performance in practical scenarios. For all real-world datasets, we standardized the features and target values to eliminate large differences in scale between them.

\subsection{Conformal Prediction Method}
\label{subsec:CPmethod}
Unlike other methods that rely on distributional assumptions (e.g., Bayesian approaches and Gaussian Processes), CP merely requires the data samples to be exchangeable \cite{Papadopoulos2011-lf}.  As full-fledged TCP can be computationally prohibitive, we are using ICP (also referred to as split CP) as described in \cite{Papadopoulos2002-bi,Lei2018-iu,Vovk2022-ar}.

ICP works as follows.  At training time, the training data is randomly split into a proper training dataset $D_{propertrain} = \left\{ (x_i, y_i) \right\}_{i=1}^{k}$ and a calibration dataset  $D_{cal} = \left\{ (x_i, y_i) \right\}_{i=k+1}^{m}$.  The regression model of choice is then trained on the proper training dataset. Subsequently, this trained model is used to provide predictions on the calibration set. Using these predictions, NCMs are determined for all points in the calibration set, which are used to construct PIs $C$. 

The prototypical choice for NCM is simply the absolute difference between the predicted output value and the true value for every point, also known as the absolute error-based NCM. Next, these values are arranged in ascending order, where we denote this ordered sequence as $\alpha_1, \ldots, \alpha_{m-k}$.  From this sequence a single index $N$ is picked to create, for any newly predicted $y$, the PIs $C = (y-\alpha_N,y+\alpha_N)$, which will have a guaranteed coverage of $1-\epsilon$.  The relevant index is determined as follows:

\begin{equation}
\label{eq:index}
    N = \left\lceil (m-k)(1-\epsilon) \right\rceil .
\end{equation}

Note that we cannot find a finite-width PI if the calibration data size $m-k-1$ is smaller than $1/\epsilon -1$, highlighting a clear limitation of ICP.

\subsection{Validity and Efficiency}
\label{subsec:validity and efficiency}

The performance of the conformal PIs at each test point $x_i$ is assessed in terms of validity and efficiency. 

We consider a PI to be valid when the fraction of intervals covering the true value over the entire test set, as shown below, reaches a predetermined coverage $1-\epsilon$. 
\begin{equation}
    \text{Validity} = \frac{1}{n_{test}} \sum_{i=1}^{n_{test}} \mathbf{1}\{y_i \in \mathcal{C}(x_i)\} .
\end{equation}
Overall, validity was assessed by repeating each experiment 100 times. The mean validity across the 100 experiments was determined and used as a performance metric.

Regarding efficiency, we determined the average length of the PIs over the entire test dataset \cite{Vovk2005-pz}, as shown below:

\begin{equation}
    \text{Efficiency} = \frac{1}{n_{test}} \sum_{i=1}^{n_{test}} |C_{i}| .
\end{equation}
Similarly,  experiments were repeated 100 times, and mean efficiency was determined and used as a performance metric. 

\subsection{Convergence}
\label{subsec:convergence}
Regarding validity, convergence is defined as the point at which the mean absolute difference between the empirical coverage rate and the target coverage rate, across two successive data sizes, is less than $1\times 10^{-3}$. For each experiment, we computed the empirical coverage rate for successive data sizes and determined the smallest data size at which this convergence condition was met.

\subsection{Models}
We used a neural network (NN), Gaussian process (GP), and quantile regression (QR) as models for the experiments. Each model offers unique advantages and potential limitations under different conditions. The models used in this paper were basic versions trained without extensive hyperparameter tuning, acknowledging that models may not always be perfectly trained or optimally tuned in practical settings. Our primary interest was to assess the quality of conformal PIs fairly, even when the model was not extensively trained, as this is a common scenario in practical applications. The code for the first two models is largely adapted from Detlefsten et al. \cite{Skafte2019-hb}.

% These models were chosen due to their broad applicability across various scientific and practical fields. 

\subsubsection{Neural Network}
For the NN, we used a mean-variance NN (MVNN), a standard model employed in various applications \cite{Nix1994-ym}. This model is designed to estimate both the mean and variance of its output distribution, assuming a certain error-distribution model. When sufficient data is available, MVNN is able to estimate uncertainty of the predictions considering the noise observed in the training data \cite{Nix1994-ym}. However, when the dataset size is insufficient, this estimation can lead to model misspecification, which can negatively impact the quality of the UQ. 

The specific NN model employed in this study consists of two separate NNs, one to predict the mean and one to predict the variance of the target variable. Each network consists of two linear layers with 50 hidden units and a sigmoid activation function. The mean prediction network provides the estimated mean, while the variance prediction network, using a softplus activation function, ensures positive variance predictions. The model was trained using the Adam optimizer, with a loss function based on a negative log-likelihood. 
% . The MVNN was chosen because one of the normalized NCM requires variance estimation. Accurate variance estimation is crucial, and the MVNN leverages the variance properties of the underlying model, eliminating the need to estimate variance by other means 
\subsubsection{Gaussian Process}
GPs naturally estimate the mean and variance of the noise distribution of the data, assuming a homoscedastic Gaussian noise distribution. When data is sparse, the model's assumptions about the distribution become more critical. The GP model utilizes a constant mean function and a radius basis function (RBF) kernel. For multi-dimensional datasets, an Automatic Relevance Determination (ARD) kernel is used to allow different length scales for each input dimension. The model is trained using the Adam optimizer, with the marginal log-likelihood as the objective function to be maximized during the optimization process \cite{Skafte2019-hb}.

\subsubsection{Quantile Regression}
QR estimates the conditional quantile, providing different quantiles of the distribution at each point. The main advantage of QR is its effectiveness in scenarios where the noise distribution is non-Gaussian and skewed, which is common in real-world applications \cite{Koenker1978-va}. Combining QR with CP, known as conformalized quantile regression (CQR), can handle varying types of noise distributions \cite{Romano2019-ci}. The QR model in this study is designed to predict multiple quantiles of the target variable. For each specified quantile, an independent gradient boosting regressor is trained with an objective function tailored to quantile regression, using the corresponding quantile value.

\subsection{Nonconformity measure}
Generally, NCMs can be divided into two major categories: absolute error-based NCMs and quantile-based NCMs \cite{Kato2023-nx}.  In this study, we investigated two commonly used absolute error-based NCMs and one quantile-based NCM, which recently started to receive more attention due to its effectiveness in handling heteroscedastic data distributions \cite{Romano2019-ci}. Although there are many other variants of NCM designed for use in combination with specific models \cite{Carlsson2014-zl, Papadopoulos2023-bj}, we chose to focus mainly on model-agnostic NCMs for their practicality. Having model-agnostic NCMs provides researchers and engineers with the flexibility to choose a model that fits the proposed experimental or practical setup.

\subsubsection{Absolute error-based nonconformity measure}
\label{subsec: Absolute error NCM}
The absolute error-based NCM is the most straightforward NCM and determines the absolute difference between the prediction $\hat{y}_i$ for $x_i$ and the true value $y_i$ and is defined as:

\begin{equation}
    \alpha_i = \mid y_i - \hat{y}_i \mid.
\end{equation}
Using this NCM, the PI is constructed as:

\begin{equation}
    \mathcal{C}(x_i) = (\hat{y}_i - \alpha_{N}, \; \hat{y}_i + \alpha_{N}),
    \label{placeholder_label}
\end{equation}
\noindent where $N$ is defined in Equation \ref{eq:index} in Subsection \ref{subsec:CPmethod}. It is important to realize that this absolute error-based NCM provides PIs that have the same length for all test examples. Consequently, this could potentially affect the efficiency of CP, especially in cases where the data exhibit varying levels of uncertainty. 
% Our aim is to investigate how this limitation may increasingly impact performance, especially with small datasets.

\subsubsection{Normalized absolute error-based nonconformity measure}
In order to overcome the possible limitation of the absolute error-based NCM described under \ref{subsec: Absolute error NCM}, a normalized absolute error-based NCM (normNCM) was developed \cite{Papadopoulos2002-bi}. The normNCM provides the possibility of having different lengths of PIs by using local weighing based on the variance $\sigma_i$ and is defined as follows:

\begin{equation}
    \alpha_i = \left| \frac{y_i - \hat{y}_i}{\sigma_i} \right|.
\end{equation}

The PIs are obtained as:

\begin{equation}
    \mathcal{C}(x_i) = (\hat{y}_i - \alpha_{N}\sigma_i, \; \hat{y}_i + \alpha_{N}\sigma_i).
    \label{placeholder_label2}
\end{equation}
In this study, $\sigma_i$ was estimated by determining the absolute residual between the predicted and true values on a logarithmic scale $\mu_i=ln(|y_i-\hat{y}_i|)$, where $\sigma_i = e^{\mu_i}$. The logarithmic scale is used to ensure that the estimate is always positive. After training the underlying
model on training data, we calculate the residuals $|y_i-\hat{y}_i|$ for all training examples $i \in D_{propertrain}$. Thereafter, the underlying model is retrained on the pairs $(x_i$, ln$(|y_i-\hat{y}_i|))$. These predicted values are used to obtain the normNCM. It should be noted that the underlying model is trained twice and these models do not necessarily have to be the same.  

%In the second approach, predicted variance using the underlying model is used to obtain $\sigma_i$. In both approaches, the quality of $\sigma_i$ highly depends on the quality of model.

\subsubsection{Quantile based-nonconformity measure}
Although the absolute error-based normNCM can provide PIs of different sizes, depending on locally estimated variances, achieving a good efficiency in homoscedatiscally distributed data is potentially problematic due to the variability caused by the estimation of the variance \cite{Lei2018-iu}. However, the effect of this variability still needs to be validated in extensive studies, as there is still limited data on the impact of its efficiency.  To overcome these potential issues, CQR has been proposed \cite{Romano2019-ci}. The main difference between the previously introduced NCMs and quantile-based NCMs is that these focus on estimating the lower and upper conditional quantiles rather than the conditional mean. During CQR, the lower and upper conditional quantile functions $q_{\epsilon_{low}}$ and $q_{\epsilon_{high}}$, are fitted to a proper training set using any quantile regression algorithm $M$ according to:

\[\{\hat{q}_{{\epsilon}_{low}},\hat{q}_{{\epsilon}_{high}}\}\leftarrow M(\{x_i, y_i\}: i \in D_{propertrain}).\]
The calibration set is then utilized to determine the conformity scores that quantify the prediction error. The resulting PI, \(\hat{C}(x) = [\hat{q}_{\text{low}}, \hat{q}_{\text{high}}]\), is assessed in the calibration set using:

\begin{equation}
    \alpha_i = \max\{\hat{q}_{\text{low}}(x_i) - y_i, y_i - \hat{q}_{\text{high}}(x_i)\}. 
\end{equation}
Subsequently, the PI for \(x_i\) is constructed as:
\begin{equation}
    C(x_i) = \left(\hat{q}_{\text{low}}(x_i) - Q_{1-\epsilon}(\alpha, D_{cal}), \hat{q}_{\text{high}}(x_i) + Q_{1-\epsilon}(\alpha, D_{cal})\right). 
\end{equation}
\noindent where $Q_{1-\epsilon}(\alpha, D_{cal})$ denotes the $(1 - \epsilon)(1 + 1/|D_{cal}|)$-th empirical quantile of $\{\alpha_i : i \in D_{cal}\}$.

\subsection{Experimental Setup}
\label{subsec:experimental_setup}
Experiments were conducted using all different models (NN, GP, QR) and NCMs (absolute error-based, normalized absolute error-based, quantile-based). The miscoverage rates (i.e., one minus the coverage rates) considered were 0.01, 0.05, 0.1, and 0.2.

For experimentation with synthetic data, we chose a noise type (homoscedastic Gaussian, heteroscedastic Gaussian, or right-skewed noise distribution) and set the noise level to 0.3.  With these settings, we then generated training sets with sizes ranging from 100 to 1000 with steps of 100.  80\% of the training data was assigned to the proper training set and the remaining 20\% constituted the calibration dataset. The test set was generated using 10,000 test samples.

For experiments with real-world data, we first selected multidimensional publicly available datasets. We then randomly subsampled the datasets to create datasets with sample sizes ranging from 100 to 1000 with steps of 100. Each dataset was split, with 80\% allocated to the training set and 20\% to the test set. The training set was then split further, with 80\% of the samples allocated for the proper training set and 20\% for the calibration dataset.

After these steps, the underlying model was trained using the proper training dataset, yielding a trained model. Subsequently, the NCM was calculated using the trained model and the calibration dataset. The trained model was then applied to the test dataset to obtain predictions. Using these predictions and the NCM, the PIs were determined. Each experiment was repeated 100 times to assess the stability of each method. Lastly, the mean validity and efficiency, as defined in Subsection \ref{subsec:validity and efficiency}, were calculated.

\section{Results}
We present the results of our experiments in two sub-sections: one for synthetic dataset (Subsection \ref{subsec: Synthetic dataset}) and one for real-world dataset (Subsection \ref{subsec: Real-world dataset}). Within both subsections, we show the performance of each model with different NCMs in terms of efficiency and validity. The procedure for calculating efficiency and validity is detailed in Subsection \ref{subsec:validity and efficiency}. A comprehensive comparison across all noise types, coverage rates, and NCMs can be found in \ref{appendix: all results}.

\subsection{Synthetic dataset}
\label{subsec: Synthetic dataset}
We present results for both one-dimensional and multi-dimensional synthetic datasets, starting with general observations regarding the relationship between validity and efficiency, followed by independent observations for each metric, before providing a more detailed analysis based on the key findings.

\begin{figure*}[htbp]
  \centering
  \includegraphics[width=\linewidth]{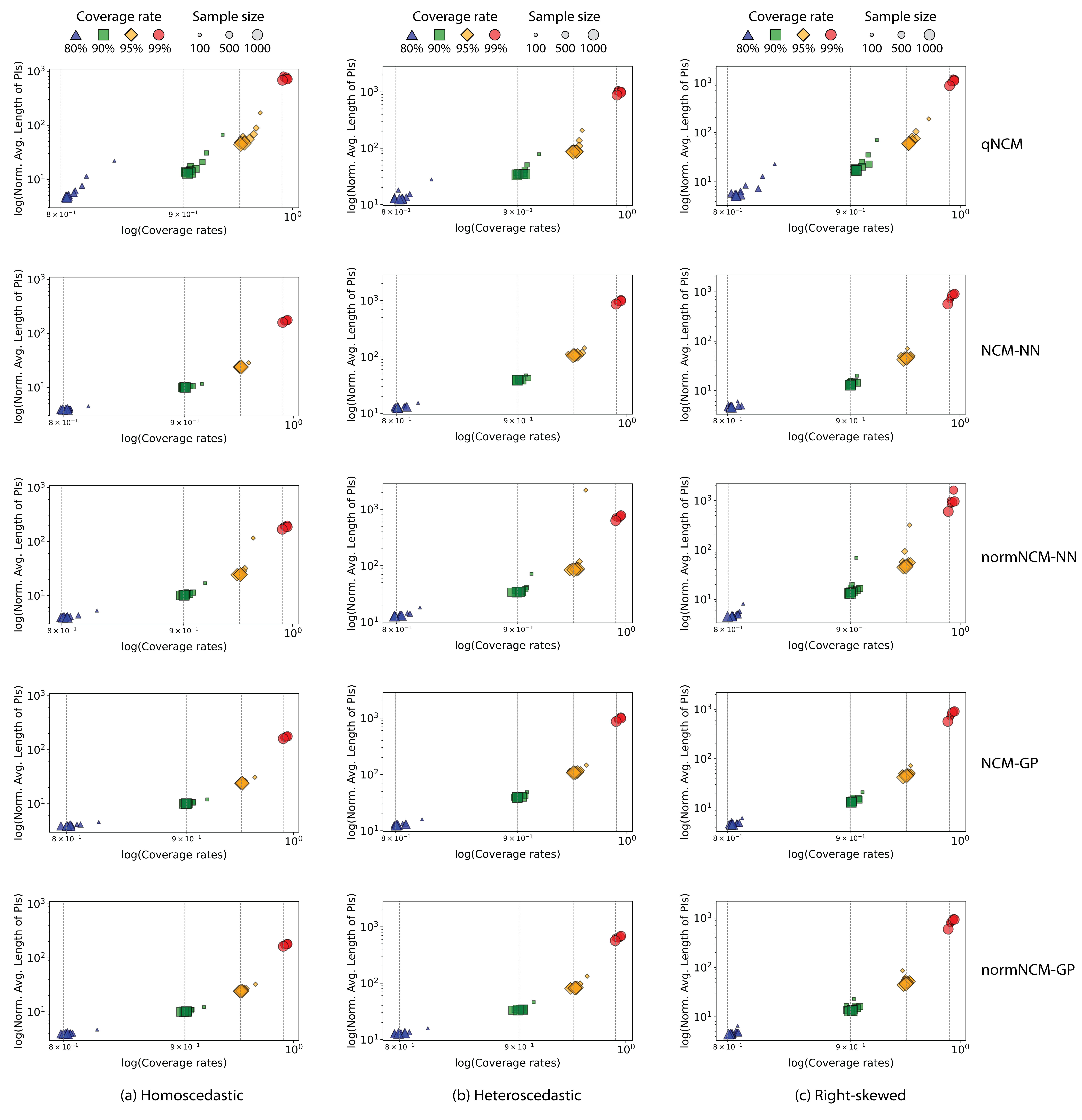}
  \caption{Trade-off between validity and normalized efficiency. In this figure, the x-axis represents validity, while the y-axis shows efficiency normalized by target validity, both plotted on a logarithmic scale. The data correspond to the one-dimensional case using homoscedastic Gaussian noise (left column) and heteroscedastic Gaussian noise (middle column) as well as right-skewed noise (right column), across varying target coverage rates and data sizes. Different shapes represent target coverage rates: triangles for 80\%, squares for 90\%, diamonds for 95\%, and circles for 99\%. The sample sizes are indicated by the size of the shapes: small for 100, medium for 500, and large for 1000. Each row corresponds to a different NCM: starting from the top, the rows correspond to qNCM, NCM-NN, normNCM-NN, NCM-GP, and normNCM-GP. 
}
  \label{fig:1d_combined}
\end{figure*}

Figure \ref{fig:1d_combined} shows the overall relationship between validity and efficiency normalized by the target coverage rates, both on a logarithmic scale, for the one-dimensional synthetic data. We observed that an increase in target coverage resulted in reduced efficiency for all NCMs and noise types, which aligns with our expectations. However, it should be noted that this trade-off varied between different models and NCMs. This is particularly evident for qNCM under homoscedastic noise conditions, as shown in the first row of Figure \ref{fig:1d_combined}. Additionally, normNCM-NN (the third row of Figure \ref{fig:1d_combined}) occasionally showed an unexpected sudden drop in efficiency for all noise types, particularly at smaller data sizes. We explore these findings in more depth in Subsection \ref{subsec:twomore}.

\begin{figure*}[htbp]
  \centering
  \includegraphics[width=0.7\linewidth]{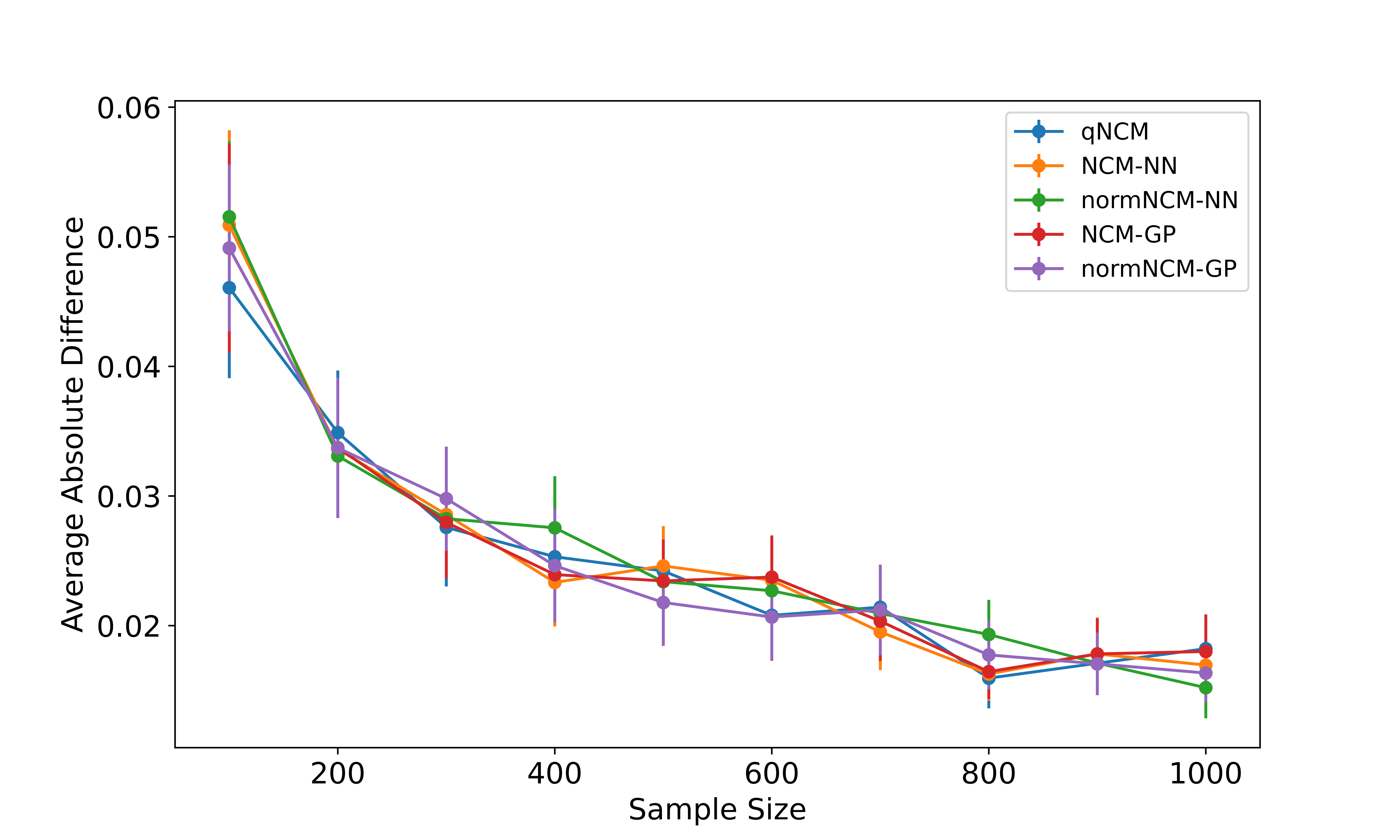}
  \caption{Mean absolute difference between empirical and target coverage rates (90\%) for different sample sizes under heteroscedastic noise conditions for different NCMs. Error bars reflect 1.96 times the SE, corresponding to 95\% confidence intervals.}
  \label{fig:90validity_heteroscedastic}
\end{figure*}

With respect to validity, the empirical coverage rate in all experiments either met or closely approximated the target coverage rate. This observation confirmed that CP guarantees validity regardless of the choice of NCM. Moreover, the minimum data size required to achieve convergence was investigated. The procedure for calculating convergence is detailed in Subsection \ref{subsec:convergence}. Overall, all NCMs exhibited approximately equal performance in terms of the datasize needed for convergence, irrespective of the target coverage rates, data size, and noise types. As an example, Figure \ref{fig:90validity_heteroscedastic} shows the mean absolute difference between the empirical coverage and the 90\% target coverage rates across various sample sizes under heteroscedastic noise conditions. The plot indicates that the mean absolute difference converges around a sample size of 800, with the standard error (SE) of the absolute difference decreasing as sample size increases, highlighting the increasing stability in validity.

With respect to efficiency, the most efficient PIs were observed under homoscedastic Gaussian noise (the first column of Figure \ref{fig:1d_combined}), whereas less efficient PIs were obtained under heteroscedastic Gaussian noise and right-skewed noise distributions (the second and the third columns of Figure \ref{fig:1d_combined}), with the exception of qNCM (the first row of Figure \ref{fig:1d_combined}
). While all NCMs exhibit increasingly less efficient PIs as the target coverage rates increase, this trend is particularly pronounced for qNCM under homoscedastic noise conditions (left in the first row of Figure \ref{fig:1d_combined}). This is summarized in Table \ref{table:efficiency comparison}, showing the mean efficiency in different data sizes for different NCMs, noise types, and target coverage rates. We observed that while data size had a limited influence, the noise type and the choice of the NCM had a more pronounced effect on the overall efficiency of the PIs. However, we did observe that the SE of the efficiency decreased as the data size increased, suggesting more stable PIs with increasing data size. 

\begin{table*}[htbp]
\small
\centering
\caption{Comparison of the mean efficiency for one-dimensional data, across different NCMs, target coverage rates, and noise types, along with the SE in parentheses.}
\label{table:efficiency comparison}
\resizebox{\linewidth}{!}{%
\begin{tabular}{c|cccc|cccc|cccc}
\hline
                                          & \multicolumn{4}{c}{\textbf{Homoscedastic}} & \multicolumn{4}{c}{\textbf{Heteroscedastic}} & \multicolumn{4}{c}{\textbf{Right-skewed}} \\ \cline{2-13} 
                                          & 80\%      & 90\%     & 95\%     & 99\%     & 80\%    & 90\%             & 95\%   & 99\%   & 80\%     & 90\%     & 95\%     & 99\%     \\ \hline
\multirow{2}{*}{\textbf{qNCM}}           & 1.53      & 2.19     & 3.39     & 7.49     & 3.04   & 4.17             & 5.41   & 10.00  & 1.71     & 2.62     & 4.10     & 10.78    \\
                                          & (0.34)    & (0.53)   & (0.61)   & (0.23)   & (0.30) & (0.44)           & (0.61) & (0.23) & (0.35)   & (0.52)   & (0.63)   & (0.47)   \\
\multirow{2}{*}{\textbf{NCM-NN}}     & 0.80      & 1.03     & 1.24     & 1.71     & 2.62   & 4.07             & 5.65   & 9.58   & 0.98     & 1.46     & 2.47     & 7.87     \\
                                          & (0.01)    & (0.02)   & (0.02)   & (0.03)   & (0.05) & (0.08)           & (0.18) & (0.17) & (0.03)   & (0.07)   & (0.13)   & (0.54)   \\
\multirow{2}{*}{\textbf{normNCM-NN}} & 0.84      & 1.12     & 1.76     & 1.88     & 2.76   & 3.95             & 15.15  & 7.23   & 1.04     & 2.07     & 4.08     & 9.99     \\
                                          & (0.03)    & (0.07)   & (0.45)   & (0.04)   & (0.10) & (0.36)           & (10.53) & (0.17)  & (0.07)   & (0.55)   & (1.34)   & (1.40)   \\
\multirow{2}{*}{\textbf{NCM-GP}}     & 0.81      & 1.04     & 1.26     & 1.71     & 2.62   & 4.10             & 5.69   & 9.60   & 1.00     & 1.51     & 2.50     & 7.88     \\
                                          & (0.01)    & (0.02)   & (0.03)   & (0.03)   & (0.07) & (0.09)           & (0.19) & (0.17) & (0.03)   & (0.08)   & (0.14)   & (0.54)   \\
\multirow{2}{*}{\textbf{normNCM-GP}} & 0.81      & 1.05     & 1.29     & 1.76     & 2.65   & 3.55             & 4.50   & 6.39   & 0.98     & 1.55     & 2.74     & 8.34     \\
                                          & (0.01)    & (0.02)   & (0.04)   & (0.03)   & (0.06) & (0.12)           & (0.25) & (0.13) & (0.04)   & (0.10)   & (0.19)   & (0.57)   \\ \hline
\end{tabular}%
}
\end{table*}

\subsubsection{Two follow-up experiments}
\label{subsec:twomore}

Based on the observations in Table \ref{table:efficiency comparison} and Figure \ref{fig:1d_combined}, we decided to perform a more detailed analysis with respect to outliers of the observed efficiencies and trends of inefficiencies of qNCM. Firstly, we focus on the observation where the normNCM-NN occasionally produced substantially less efficient PIs across various noise types and target coverage rates, particularly with smaller datasets (the third row of Figure \ref{fig:1d_combined}). In order to quantify the effect of sudden less efficient PIs, we defined experimental outliers if the SE (on a logarithmic scale) exceeded the threshold defined by the interquartile range method with a multiplier of 1.5.

To determine the fraction of outliers, we further analyzed the 100 repetitions within each experiment to determine how many repetitions were themselves considered outliers. Using this definition, normNCM-NN was the only NCM showing outliers. Based on the identified outliers, a corrected efficiency and SE were determined by omitting the experimental outliers. Table \ref{table:outliers} shows the detailed information of the identified outliers in each experiment, and corrected efficiency and SE after removing the outliers. 

\begin{table*}[htbp]
\small
\caption{Comparison of efficiency and corrected efficiency with SE in parentheses, and outlier percentages for the identified outliers.}
\label{table:outliers}
\resizebox{\linewidth}{!}{%

\begin{tabular}{c|c|c|ccc}
\hline
                              & \textbf{Homoscedastic} & \textbf{Heteroscedastic} & \multicolumn{3}{c}{\textbf{Right-skewed}}           \\ \hline
Target coverage rate        & 95\%                   & 95\%                     & 90\%            & 95\%            & 99\%            \\
NCM                           & normNCM-NN        & normNCM-NN          & normNCM-NN & normNCM-NN & normNCM-NN \\
Data size                     & 100                    & 100                      & 100             & 100             & 700             \\
Efficiency {[}SE{]}           & 5.82{[}2.08{]}         & 109.96{[}91.40{]}        & 6.93 {[}3.75{]} & 15.94{[}5.91{]} & 16.32{[}6.48{]} \\
Corrected Efficiency {[}SE{]} & 1.84{[}0.01{]}         & 5.11{[}0.02{]}           & 3.17{[}0.03{]}  & 3.20{[}0.03{]}  & 3.23{[}0.03{]}  \\
Outliers {[}\%{]}             & 9                      & 11                       & 3               & 1               & 1               \\ \hline
\end{tabular}%
}
\end{table*}

The overall SE decreased substantially, with an average reduction of 89.32\% after omitting the outliers, leading to more stable results. Although no direct clear pattern could be observed, the number of outliers for the normNCM-NN was more pronounced for the 95\% target coverage rate for both homoscedastic and heteroscedastic Gaussian noise at a data size of 100 samples. On average, 5 repetitions per experiment were considered to be outliers.

\begin{figure*}[htbp] 
  \centering 
  \begin{subfigure}[b]{0.49\linewidth} 
  \centering  % Centers the image within the subfigure
  \includegraphics[width=\linewidth]{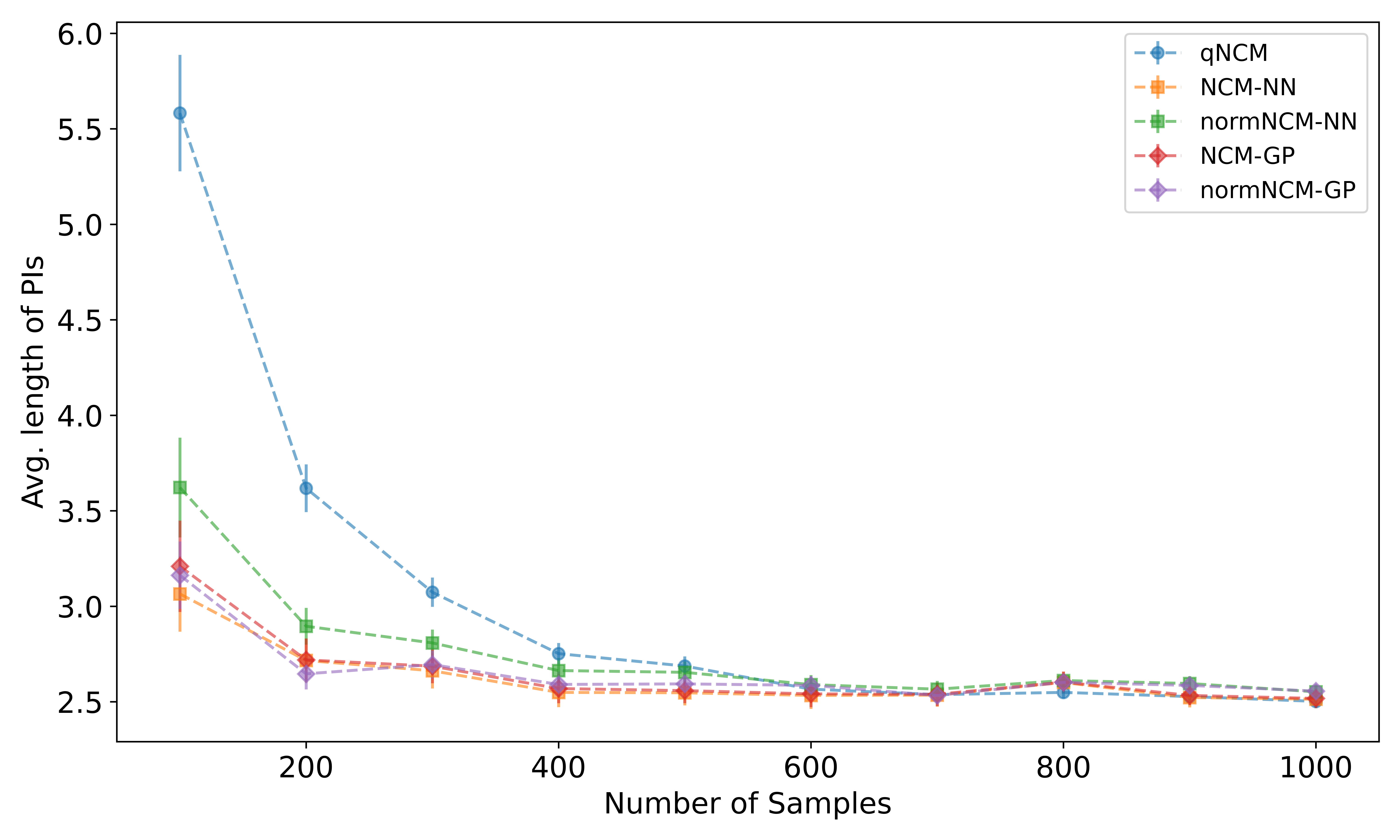}
  \caption{Gaussian heteroscedastic}
  \label{fig:Gaussian heteroscedastic}
  \end{subfigure} 
  \hfill 
  \begin{subfigure}[b]{0.49\linewidth} 
  \centering  % Centers the image within the subfigure
  \includegraphics[width=\linewidth]{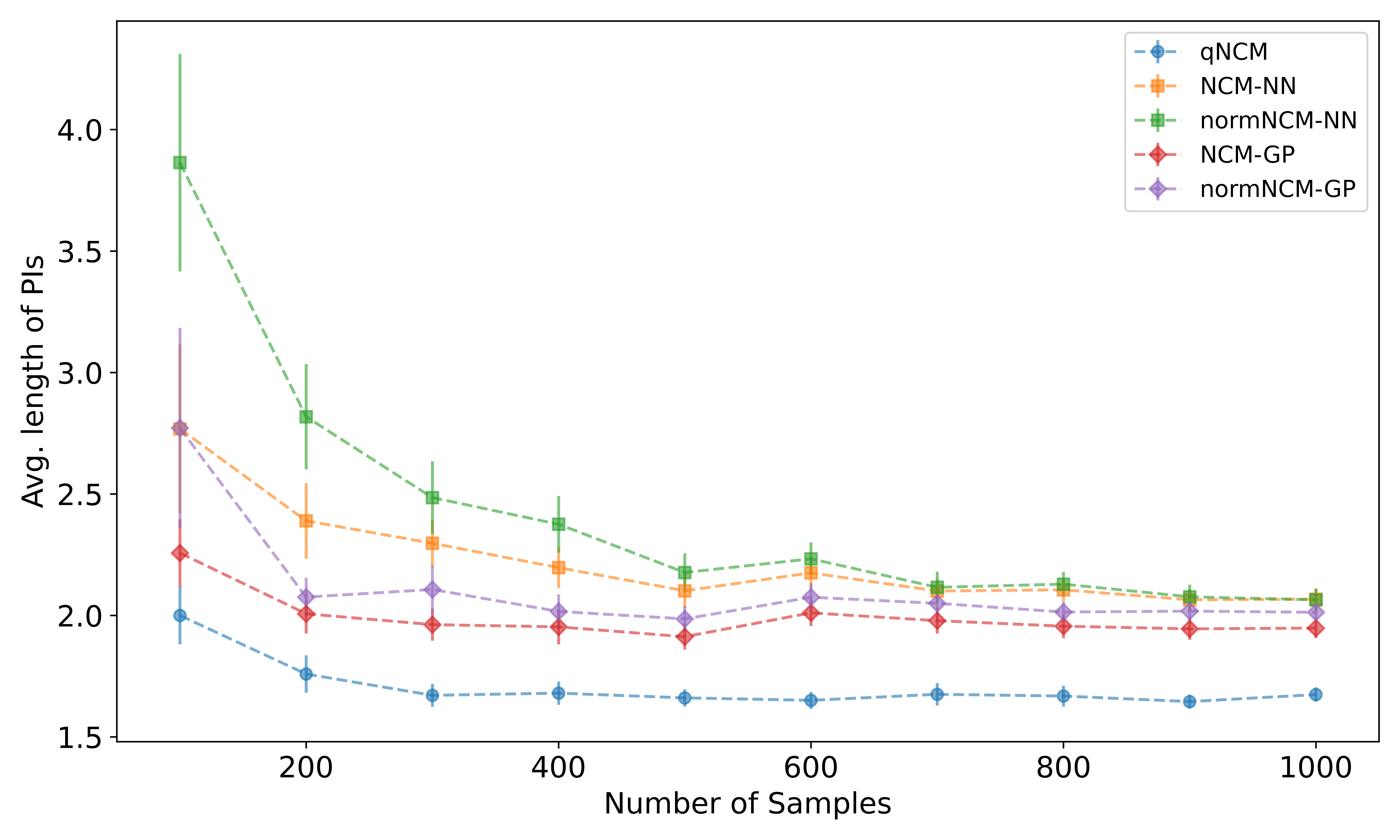}
  \caption{Non-Gaussian heteroscedastic}
  \label{fig:Non-Gaussian heteroscedastic}
  \end{subfigure} 
  \caption{Efficiency for varying data sizes under (a) heteroscedastic Gaussian and (b) heteroscedastic Non-Gaussian noise conditions. The plot shows the performance of different NCMs
(qNCM, NCM-NN, normNCM-NN, NCM-GP and normNCM-GP) for an 80\% target coverage rate, with error bars reflecting 1.96 times the SE, corresponding to
95\% confidence intervals.}
  \label{fig:qNCM-comparison} 
\end{figure*}

To investigate the inefficiency tendencies of qNCM under heteroscedastic noise (the second column of Figure \ref{fig:1d_combined}), given that qNCM is specifically designed to adapt to heteroscedasticity in data \cite{Romano2019-ci}, another additional experiment was conducted. It is important to note that this inefficiency occurs primarily with smaller data sizes and smaller SEs compared to other NCMs, indicating that this is a stable result. This effect is shown in Figure \ref{fig:Gaussian heteroscedastic}. In this figure, both qNCM and normNCM-NN showed lower efficiency compared to other NCMs at a data size of 100. 
% However, a notable difference is that the SE of qNCM is 77.62\% smaller than that of normNCM-NN, suggesting better stability despite the inefficiency.

\begin{figure*}[htbp]
\centering
  \includegraphics[width=\linewidth]{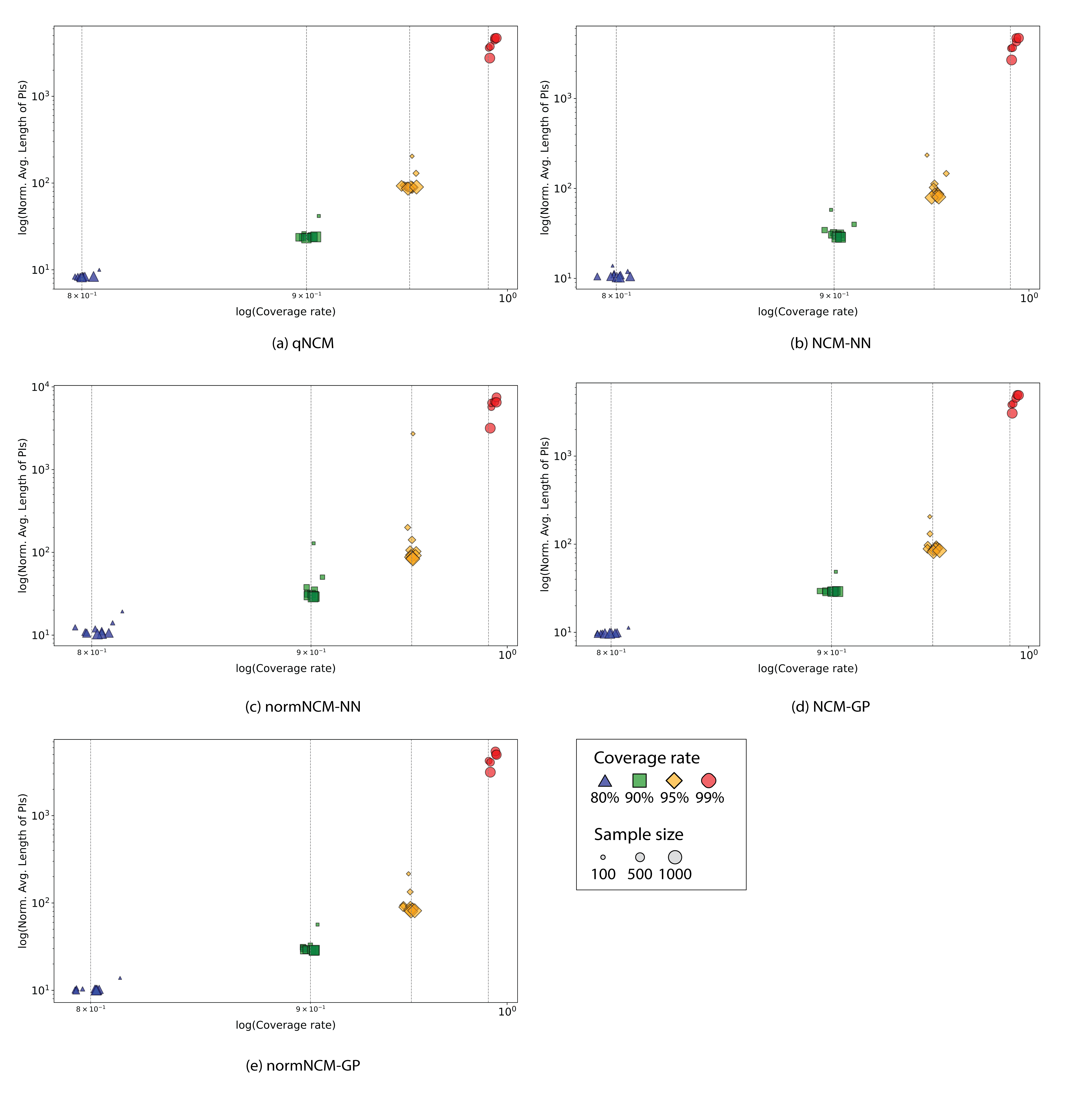}
    \caption{Trade-off between validity and normalized efficiency. In this figure, the x-axis represents validity, while the y-axis shows efficiency normalized by target validity, both plotted on a logarithmic scale. The data correspond to the one-dimensional case using non-Gaussian heteroscedastic noise across varying target coverage rates and data sizes. Similar to Figure \ref{fig:1d_combined}, different shapes and sizes represent different target coverage rates and sample sizes. The different NCMs used are: a) qNCM, b) NCM-NN, c) normNCM-NN, d) NCM-GP, and e) normNCM-GP.}
    \label{fig:heteroscedastic_nonGaussian}
\end{figure*}

In the additional experiment, we used heteroscedastic non-Gaussian noise dataset (with outliers) from \cite{Romano2019-ci} to further explore the differences in efficiency for this particular noise type. Table \ref{table:heteroNonGaussian-efficiency} presents a comparison of the mean efficiency, averaged over the different data sizes, across different NCMs and target coverage rates under heteroscedastic non-Gaussian noise. The results indicated that qNCM performed better under this noise type, as shown in Figure \ref{fig:heteroscedastic_nonGaussian}. This improved performance was noticeable even with smaller data sizes, as shown in Figure \ref{fig:Non-Gaussian heteroscedastic}, where qNCM demonstrated the best efficiency along with smaller SEs. Furthermore, as shown in this figure, it seems that increasing the dataset size does not consistently lead to improved efficiency, and this trend holds across different noise types under different conditions.

\begin{table*}[htbp]
\small
\centering
\caption{Comparison of the mean efficiency for one-dimensional data under heteroscedastic non-Gaussian noise, across different NCMs and target coverage rates along with the SE in parentheses.}
\label{table:heteroNonGaussian-efficiency}
% \resizebox{\textwidth}{!}{%

\begin{tabular}{c|cccc}
\hline
                                          & \textbf{80\%} & \textbf{90\%} & \textbf{95\%} & \textbf{99\%} \\ \hline
\multirow{2}{*}{\textbf{qNCM}}           & 1.71  & 2.58  & 5.28  & 40.03  \\
                                          & (0.03) & (0.18) & (0.58) & (3.13) \\
\multirow{2}{*}{\textbf{NCM-NN}}     & 2.23  & 3.43  & 5.51  & 39.38  \\
                                          & (0.07)  & (0.28) & (0.76) & (3.19) \\
\multirow{2}{*}{\textbf{normNCM-NN}} & 2.43  & 4.33  & 18.52 & 59.77  \\
                                          & (0.18)  & (0.97) & (13.02) & (6.07) \\
\multirow{2}{*}{\textbf{NCM-GP}}     & 1.99  & 3.13  & 5.24  & 42.07  \\
                                          & (0.03)  & (0.19) & (0.60)  & (2.99) \\
\multirow{2}{*}{\textbf{normNCM-GP}} & 2.11  & 3.24  & 5.20  & 44.89  \\
                                          & (0.07)  & (0.27) & (0.67)  & (3.41) \\ \hline
\end{tabular}
\end{table*}

\begin{figure*}[htbp]
\centering
  \includegraphics[width=\linewidth]{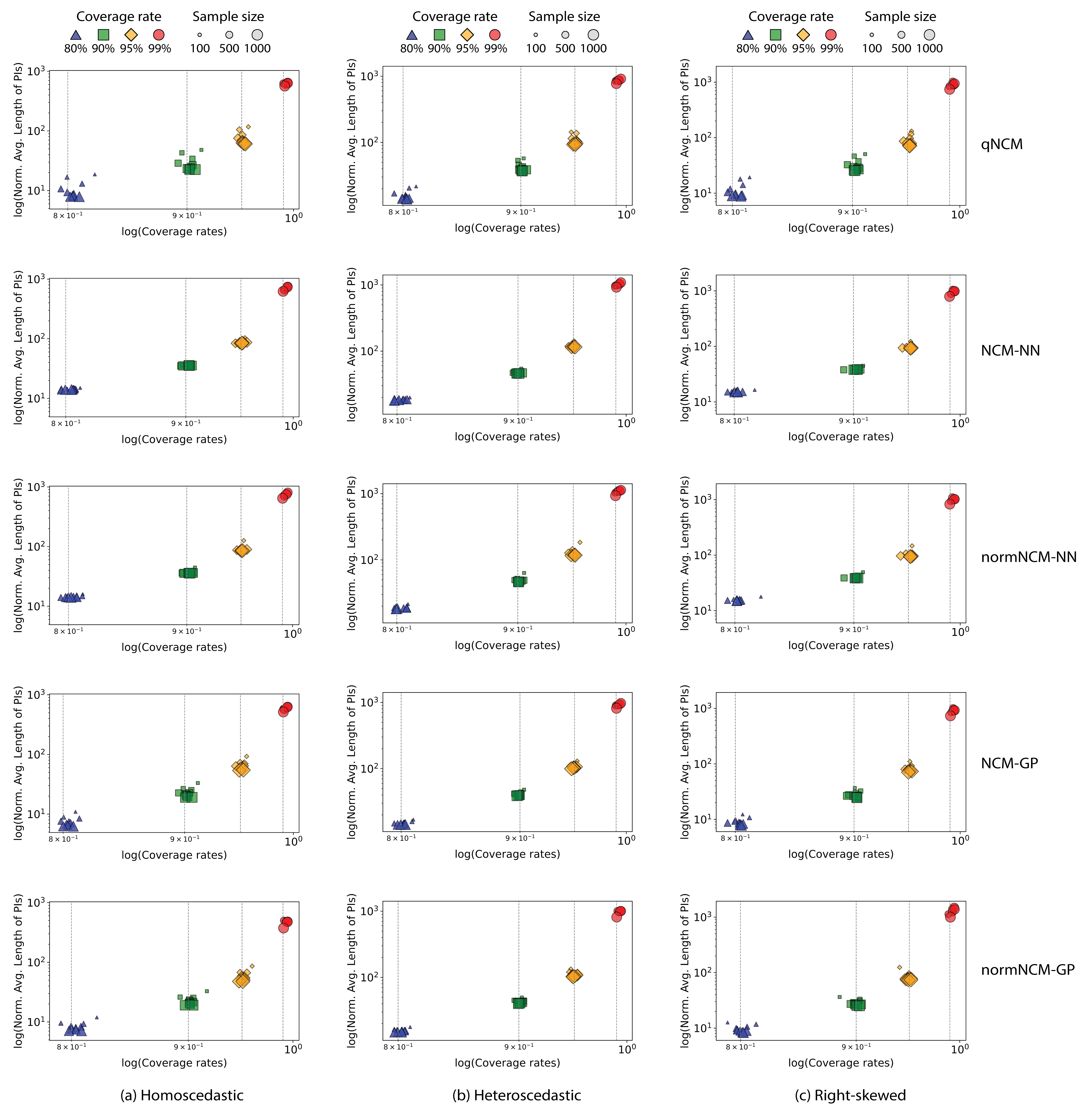}
  \caption{Trade-off between validity and normalized efficiency. In this figure, the x-axis represents validity, while the y-axis shows efficiency normalized by validity, both plotted on a logarithmic scale. The data correspond to the five-dimensional case using homoscedastic Gaussian noise (right column), heteroscedastic Gaussian noise (middle column), and right-skewed noise (left column), across varying target coverage rates and data sizes. Similar to Figure \ref{fig:1d_combined}, different shapes and sizes represent different target coverage rates and sample sizes. Each row corresponds to a different NCM: starting from
the top, the rows correspond to qNCM, NCM-NN, normNCM-NN, NCM-GP, and normNCM-GP.
}
  \label{fig:5dcombined}
\end{figure*}

When comparing the results of multi-dimensional datasets, a similar trend was observed as with the one-dimensional datasets. Figure \ref{fig:5dcombined} shows the relationship between validity and efficiency for five-dimensional synthetic data. A trade-off between validity and efficiency normalized by the target coverage rates was noted across all dimensionalities, noise types, target rates, and NCMs. 

%Overall, NCM-NN and normNCM-NN showed generally less efficient PIs compared to other NCMs, while qNCM, NCM-GP, and normNCM-GP demonstrated greater variability in efficiency, especially with smaller data. This was most evident under homoscedastic noise for all dimensionalities ($d = 2, 5,$ and $8$).

The validity of all experiments either met or closely approached the target coverage rates, with a decrease in the SE of validity as the data size increased. Similar to the one-dimensional datasets, we examined the minimum data size required to achieve convergence in validity. Overall, all NCMs performed similarly, as observed with one-dimensional data, and increasing the dimensionality did not necessarily negatively impact the minimum data size required for convergence. 

In terms of efficiency, a gradual decrease, along with larger SEs, was observed as the dimensionality increased. These trends are shown in Table \ref{table:efficiency comparison-2d} for two-dimensional data, Table \ref{table:efficiency comparison-5d} for five-dimensional data, and Table \ref{table:efficiency comparison-8d} for eight-dimensional data.

\begin{table*}[htbp]
\small % To make the content size smaller
\caption{Comparison of the mean efficiency for two-dimensional data, across different NCMs, target coverage rates, and noise types, along with the SE in parentheses.}
\label{table:efficiency comparison-2d}
\resizebox{\linewidth}{!}{%

\begin{tabular}{c|cccc|cccc|cccc}
\hline
                                          & \multicolumn{4}{c}{\textbf{Homoscedastic}} & \multicolumn{4}{c}{\textbf{Heteroscedastic}} & \multicolumn{4}{c}{\textbf{Right-skewed}} \\ \cline{2-13} 
                                          & 80\%      & 90\%     & 95\%     & 99\%     & 80\%    & 90\%             & 95\%   & 99\%   & 80\%     & 90\%     & 95\%     & 99\%     \\ \hline
\multirow{2}{*}{\textbf{qNCM}}           & 1.31      & 1.70     & 2.10     & 2.04     & 1.64    & 2.18             & 2.73   & 2.66   & 1.44     & 2.05     & 2.99     & 4.62     \\
                                          & (0.15)    & (0.20)   & (0.23)   & (0.56)   & (0.13)  & (0.18)           & (0.23) & (0.73) & (0.17)   & (0.19)   & (0.22)   & (1.29)   \\
\multirow{2}{*}{\textbf{NCM-NN}}     & 1.35      & 1.76     & 2.24     & 2.69     & 1.67    & 2.29             & 3.09   & 3.49   & 1.56     & 2.08     & 3.15     & 5.05     \\
                                          & (0.01)    & (0.01)   & (0.06)   & (0.74)   & (0.02)  & (0.05)           & (0.09) & (0.96) & (0.01)   & (0.04)   & (0.11)   & (1.40)   \\
\multirow{2}{*}{\textbf{normNCM-NN}} & 1.35      & 1.78     & 2.26     & 2.79     & 1.69    & 2.31             & 3.13   & 3.51   & 1.57     & 2.11     & 3.22     & 5.13     \\
                                          & (0.01)    & (0.02)   & (0.07)   & (0.77)   & (0.02)  & (0.05)           & (0.11) & (0.96) & (0.01)   & (0.05)   & (0.14)   & (1.43)   \\
\multirow{2}{*}{\textbf{NCM-GP}}     & 0.85      & 1.12     & 1.41     & 1.46     & 1.36    & 1.92             & 2.53   & 2.57   & 1.05     & 1.62     & 2.72     & 4.77     \\
                                          & (0.02)    & (0.03)   & (0.07)   & (0.40)   & (0.03)  & (0.05)           & (0.09) & (0.70) & (0.04)   & (0.08)   & (0.14)   & (1.33)   \\
\multirow{2}{*}{\textbf{normNCM-GP}} & 0.90      & 1.19     & 1.49     & 1.29     & 1.41    & 1.90             & 2.38   & 2.12   & 1.06     & 1.77     & 3.20     & 5.66     \\
                                          & (0.03)    & (0.04)   & (0.07)   & (0.35)   & (0.03)  & (0.07)           & (0.11) & (0.58) & (0.05)   & (0.11)   & (0.30)   & (1.61)   \\ \hline
\end{tabular}%
}
\end{table*}

\begin{table*}[htbp]
\small % To make the content size smaller
\caption{Comparison of the mean efficiency for five-dimensional data, across different NCMs, target coverage rates, and noise types, along with the SE in parentheses.}
\label{table:efficiency comparison-5d}
\resizebox{\linewidth}{!}{%

\begin{tabular}{c|cccc|cccc|cccc}
\hline
                                          & \multicolumn{4}{c}{\textbf{Homoscedastic}} & \multicolumn{4}{c}{\textbf{Heteroscedastic}} & \multicolumn{4}{c}{\textbf{Right-skewed}} \\ \cline{2-13} 
                                          & 80\%      & 90\%     & 95\%     & 99\%     & 80\%    & 90\%             & 95\%   & 99\%   & 80\%     & 90\%     & 95\%     & 99\%     \\ \hline
\multirow{2}{*}{\textbf{qNCM}}           & 2.19      & 2.96     & 3.83     & 3.71     & 3.22    & 4.29             & 5.40   & 5.15   & 2.39     & 3.30     & 4.41     & 5.35     \\
                                          & (0.25)    & (0.29)   & (0.32)   & (1.01)   & (0.18)  & (0.23)           & (0.29) & (1.41) & (0.25)   & (0.28)   & (0.32)   & (1.47)   \\
\multirow{2}{*}{\textbf{NCM-NN}}     & 2.84      & 3.61     & 4.38     & 4.16     & 3.70    & 4.77             & 6.07   & 6.07   & 3.04     & 3.93     & 4.88     & 5.65     \\
                                          & (0.02)    & (0.04)   & (0.09)   & (1.14)   & (0.03)  & (0.08)           & (0.14) & (1.66) & (0.03)   & (0.07)   & (0.14)   & (1.55)   \\
\multirow{2}{*}{\textbf{normNCM-NN}} & 2.89      & 3.72     & 4.60     & 4.43     & 3.79    & 4.98             & 6.49   & 6.40   & 3.12     & 4.04     & 5.17     & 5.89     \\
                                          & (0.04)    & (0.08)   & (0.20)   & (1.21)   & (0.06)  & (0.16)           & (0.32) & (1.75) & (0.05)   & (0.10)   & (0.25)   & (1.62)   \\
\multirow{2}{*}{\textbf{NCM-GP}}     & 1.52      & 2.30     & 3.25     & 3.54     & 2.98    & 4.15             & 5.44   & 5.48   & 1.83     & 2.81     & 4.09     & 5.34     \\
                                          & (0.09)    & (0.14)   & (0.18)   & (0.97)   & (0.06)  & (0.09)           & (0.14) & (1.50) & (0.09)   & (0.12)   & (0.19)   & (1.47)   \\
\multirow{2}{*}{\textbf{normNCM-GP}} & 1.69      & 2.30     & 2.95     & 2.80     & 3.14    & 4.31             & 5.57   & 5.81   & 1.96     & 2.87     & 4.24     & 7.63     \\
                                          & (0.09)    & (0.13)   & (0.19)   & (0.77)   & (0.06)  & (0.09)           & (0.15) & (1.59) & (0.09)   & (0.12)   & (0.25)   & (2.12)   \\ \hline
\end{tabular}%
}
\end{table*}

\begin{table*}[htbp]
\small % To make the content size smaller
\caption{Comparison of the mean efficiency for eight-dimensional data, across different NCMs, target coverage rates, and noise types, along with the SE in parentheses.}
\label{table:efficiency comparison-8d}
\resizebox{\linewidth}{!}{%

\begin{tabular}{c|cccc|cccc|cccc}
\hline
                                          & \multicolumn{4}{c}{\textbf{Homoscedastic}} & \multicolumn{4}{c}{\textbf{Heteroscedastic}} & \multicolumn{4}{c}{\textbf{Right-skewed}} \\ \cline{2-13} 
                                          & 80\%      & 90\%     & 95\%     & 99\%     & 80\%    & 90\%             & 95\%   & 99\%   & 80\%     & 90\%     & 95\%     & 99\%     \\ \hline
\multirow{2}{*}{\textbf{qNCM}}           & 3.10      & 4.19     & 5.33     & 8.18     & 4.89    & 6.45             & 7.80   & 7.39   & 3.33     & 4.51     & 5.82     & 6.04     \\
                                          & (0.24)    & (0.28)   & (0.30)   & (0.15)   & (0.18)  & (0.23)           & (0.30) & (2.02) & (0.25)   & (0.29)   & (0.33)   & (1.65)   \\
\multirow{2}{*}{\textbf{NCM-NN}}     & 3.97      & 5.04     & 6.06     & 8.66     & 5.53    & 7.11             & 8.78   & 8.48   & 4.17     & 5.33     & 6.51     & 6.32     \\
                                          & (0.05)    & (0.07)   & (0.13)   & (0.18)   & (0.08)  & (0.15)           & (0.25) & (2.32) & (0.06)   & (0.09)   & (0.18)   & (1.73)   \\
\multirow{2}{*}{\textbf{normNCM-NN}} & 4.13      & 5.40     & 6.71     & 9.59     & 5.78    & 7.67             & 9.90   & 9.24   & 4.34     & 5.68     & 7.14     & 6.82     \\
                                          & (0.11)    & (0.21)   & (0.41)   & (0.24)   & (0.20)  & (0.40)           & (0.79) & (2.52) & (0.12)   & (0.23)   & (0.41)   & (1.86)   \\
\multirow{2}{*}{\textbf{NCM-GP}}     & 2.63      & 3.74     & 4.96     & 8.08     & 4.69    & 6.38             & 8.12   & 8.04   & 2.86     & 4.11     & 5.56     & 6.13     \\
                                          & (0.13)    & (0.11)   & (0.13)   & (0.21)   & (0.08)  & (0.11)           & (0.21) & (2.20) & (0.13)   & (0.14)   & (0.18)   & (1.68)   \\
\multirow{2}{*}{\textbf{normNCM-GP}} & 2.74      & 3.58     & 4.43     & 6.86     & 4.79    & 6.41             & 8.06   & 7.79   & 2.98     & 3.98     & 5.12     & 6.54     \\
                                          & (0.12)    & (0.14)   & (0.18)   & (0.24)   & (0.09)  & (0.13)           & (0.20) & (2.13) & (0.13)   & (0.15)   & (0.23)   & (1.80)   \\ \hline
\end{tabular}%
}
\end{table*}

While the normNCM-NN occasionally produced inefficient PIs under heteroscedastic noise in one-dimensional data, these occurrences were not observed in higher dimensions. In fact, no outliers were detected from normNCM-NN. Instead, certain conditions using normNCM-GP with relatively large datasets were identified as outliers under right-skewed noise. Detailed information on the outliers detected in each experiment and the corrected efficiency and SE after removing the outliers are presented in Table \ref{table:outliers-hd}. Omitting outliers led to an improvement in the efficiency of the PIs and more stable results as indicated by the reduced SE. 

\begin{table*}[htbp]
\small
\centering
\caption{Comparison of efficiency, corrected efficiency with SE in parentheses, and outlier percentages for the identified outliers for multi-dimensional dataset.}
\label{table:outliers-hd}

\begin{tabular}{c|ccc}
\hline
                              & \multicolumn{3}{c}{\textbf{Right-skewed}}           \\ \cline{2-4} 
                              & $d=5$            & $d=8$            & $d=8$            \\ \hline
Target coverage rate        & 99\%             & 99\%             & 99\%             \\
NCM                           & normNCM-GP       & normNCM-GP       & normNCM-GP       \\
Data size                     & 800              & 800              & 900              \\
Efficiency {[}SE{]}           & 14.88 {[}1.05{]} & 11.57 {[}0.73{]} & 12.67 {[}0.70{]} \\
Corrected Efficiency {[}SE{]} & 3.15 {[}0.02{]}  & 5.72 {[}0.02{]}  & 5.74 {[}0.02{]}  \\
Outliers {[}\%{]}             & 5                & 5                & 1                \\ \hline
\end{tabular}
\end{table*}

\subsection{Real-world dataset}
\label{subsec: Real-world dataset}
\begin{figure*}[htbp]
  \centering
  \includegraphics[width=\linewidth]{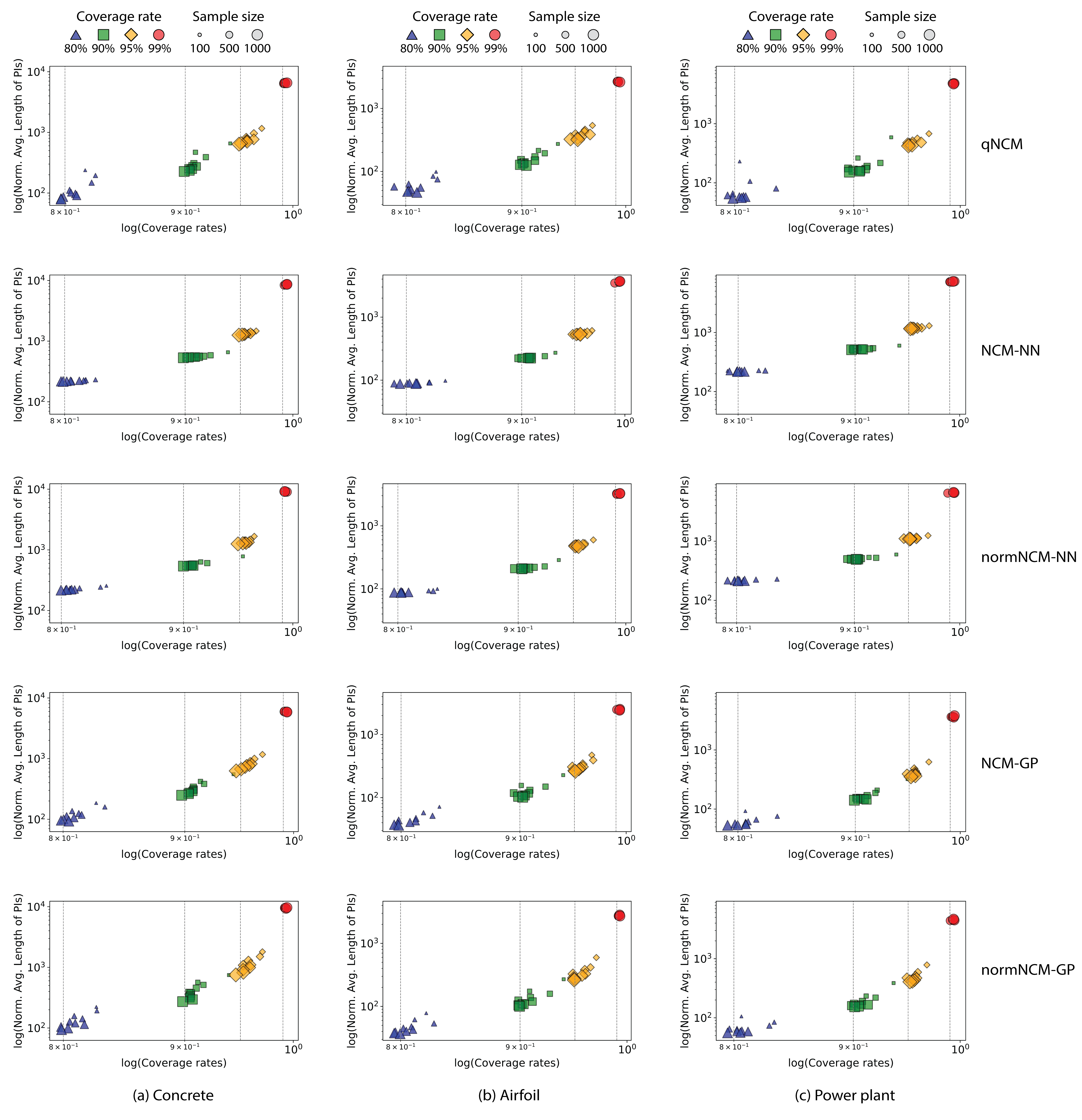}
  \caption{Trade-off between validity and normalized efficiency. In this figure, the x-axis represents validity, while the y-axis shows efficiency normalized by validity, both plotted on a logarithmic scale. The data correspond to the real-world data case using Concrete Compressive Strength (left column) and Airfoil Self-Noise (middle column) as well as Combined Cycle Power Plant (right column), across varying target coverage rates and data sizes. Similar to Figure \ref{fig:1d_combined}, different shapes and sizes represent different target coverage rates and sample sizes. Each row corresponds to a different NCM: starting from the top, the rows correspond to qNCM, NCM-NN, normNCM-NN, NCM-GP, and normNCM-GP.
}
  \label{fig:combined-realworld}
\end{figure*}

When analyzing the results of real-world datasets, a trend similar to that observed in synthetic datasets emerged. Figure \ref{fig:combined-realworld} shows the relationship between validity and efficiency for real-world data. A trade-off between validity and efficiency normalized by the target coverage rates was noted across all datasets, noise types, target coverage rates, and NCMs. From this figure, it is observed that NCM-NN and normNCM-NN consistently produced less efficient PIs compared to other NCMs (the second and the third rows of Figure \ref{fig:combined-realworld}). Furthermore, qNCM, NCM-GP, and normNCM-GP showed greater variability in efficiency, particularly as data size decreased. For all NCMs, especially at the 95\%, 90\%, and 80\% target coverage rates, smaller datasets tended to produce more conservative validity.

With respect to validity, similar to the synthetic datasets, in nearly all experiments, the empirical coverage rate either met or closely approached the target coverage rate on average, reconfirming that CP ensures the promised validity regardless of the choice of NCMs. The SE of validity decreased as the data size increased. As with the synthetic datasets, we analyzed the minimum data size needed for validity convergence in the real-world datasets. The results showed consistent performance across all NCMs, with no noticeable impact from using the different dataset. 

Regarding efficiency, Table \ref{table:efficiecny comparison-realworld} presents the mean efficiency, averaged over the different data sizes, across different NCMs and target coverage rates and datasets. Overall, NCM-NN and normNCM-NN showed less efficient PIs than the other NCMs. Increasing the data size did not lead to significant improvements in efficiency for these two NCMs across all datasets, as observed in the synthetic data. We conducted the same outlier detection analysis, but no points were identified as outliers.

\begin{table*}[htbp]
\small % To make the content size smaller
\caption{Comparison of the mean efficiency for real-world data, across different NCMs, target coverage rates and noise types, along with
the SE in parentheses.}
\label{table:efficiecny comparison-realworld}
\resizebox{\linewidth}{!}{%

\begin{tabular}{c|cccc|cccc|cccc}
\hline
                                          & \multicolumn{4}{c}{\textbf{Concrete} {[}$d = 9${]}} & \multicolumn{4}{c}{\textbf{Airfoil} {[}$d = 6${]}} & \multicolumn{4}{c}{\textbf{Power plant} {[}$d = 4${]}} \\ \cline{2-13} 
                                          & 80\%      & 90\%     & 95\%     & 99\%     & 80\%    & 90\%             & 95\%   & 99\%   & 80\%     & 90\%     & 95\%     & 99\%     \\ \hline
\multirow{2}{*}{\textbf{q-NCM}}           & 24.56     & 33.43    & 36.00    & 26.00    & 12.56   & 16.75            & 17.59  & 10.59  & 16.55    & 22.38    & 22.21    & 19.22    \\
                                          & (3.43)    & (4.33)   & (4.71)   & (10.62)  & (1.11)  & (1.50)           & (2.23) & (4.32) & (3.43)   & (4.20)   & (2.75)   & (7.85)   \\
\multirow{2}{*}{\textbf{NCM-NN}}     & 44.58     & 56.33    & 59.63    & 34.21    & 18.09   & 23.26            & 24.70  & 14.46  & 44.21    & 53.07    & 53.79    & 29.01    \\
                                          & (0.40)    & (1.19)   & (6.70)   & (13.97)  & (0.20)  & (0.47)           & (2.77) & (5.91) & (0.24)   & (0.83)   & (6.01)   & (11.85)  \\
\multirow{2}{*}{\textbf{normNCM-NN}} & 45.45     & 58.99    & 61.57    & 36.44    & 18.07   & 22.27            & 22.61  & 12.96  & 43.69    & 51.41    & 50.62    & 26.43    \\
                                          & (0.81)    & (2.32)   & (7.11)   & (14.88)  & (0.25)  & (0.75)           & (2.58) & (5.29) & (0.38)   & (1.00)   & (5.67)   & (10.79)  \\
\multirow{2}{*}{\textbf{NCM-GP}}     & 24.62     & 33.87    & 37.26    & 23.70    & 9.31    & 13.22            & 14.60  & 9.92   & 12.58    & 17.78    & 19.04    & 14.62    \\
                                          & (1.89)    & (2.93)   & (4.84)   & (9.68)   & (0.70)  & (1.21)           & (1.92) & (4.05) & (0.79)   & (1.73)   & (2.48)   & (5.97)   \\
\multirow{2}{*}{\textbf{normNCM-GP}} & 27.46     & 42.58    & 51.34    & 38.13    & 9.62    & 14.06            & 15.87  & 10.99  & 13.62    & 20.08    & 22.64    & 18.09    \\
                                          & (2.68)    & (4.67)   & (7.64)   & (15.57)  & (0.83)  & (1.62)           & (2.35) & (4.49) & (1.01)   & (2.24)   & (3.09)   & (7.34)   \\ \hline
\end{tabular}%
}
\end{table*}

\section{Discussion}
Our study systematically evaluated various NCMs using both synthetic and real-world datasets under different conditions. The results highlight that, as expected, ICP with different NCMs consistently maintains its validity across all experimental settings. However, efficiency was significantly influenced by the choice of NCMs, largely due to each NCM’s inherent sensitivity to data characteristics. This variability makes it difficult to propose a definitive guideline for selecting the best NCM across all scenarios.

The type of noise seems to have the most pronounced impact, irrespective of the kind of NCM employed. Particularly regarding qNCM, our experiments based on synthetic data revealed that its performance can be highly dependent on the type of heteroscedasticity present in the data. The outstanding performance of qNCM in heteroscedastic non-Gaussian noise aligns with its ability to adapt to such noise distributions, which is evident even with small datasets. Combined with qNCM estimating its PI's lower and upper quantiles separately \cite{Romano2019-ci}, this makes qNCM particularly effective when dealing with data exhibiting asymmetric noise distributions (Figure \ref{fig:Non-Gaussian heteroscedastic}).  It should be noted, however, that the separate modeling of a PI's lower and upper range is more a modeling choice than an inherent property of the NCM.  In fact, all NCMs considered in our study can also be adapted to deal with asymmetric intervals.

% , even more so in scenarios where outliers are present
On the other hand, modeling asymmetric PIs can result in a loss of efficiency when applied to symmetric noise types, such as homoscedastic Gaussian or heteroscedastic Gaussian noise as it can lead to unnecessarily asymmetric PIs, for example, as shown in Table \ref{table:efficiency comparison} and Figure \ref{fig:Gaussian heteroscedastic}. This issue is particularly pronounced in smaller datasets, where qNCM may easily underestimate the symmetry in the noise distribution, resulting in less efficient PIs. The inefficiency of qNCM under homoscedastic noise, where the noise distribution is symmetric and uniform, further illustrates this limitation.

% Contrary to qNCM, NCM-GP, and NCM-NN showed increasing inefficiency when heteroscedasticity was present in the data, which aligns with the known properties of these NCMs \cite{Papadopoulos2002-bi}.  B
Furthermore, both normNCM-NN and normNCM-GP, designed to flexibly adapt to heteroscedastic noise, produced slightly more inefficient PIs under homoscedastic noise compared to other NCMs, for instance, as shown in Table \ref{table:efficiency comparison}. This behavior was also noted in \cite{Lei2018-iu}, and our study demonstrates that this inefficiency becomes even more pronounced for smaller data sizes.

Notably, the aforementioned normNCMs occasionally produced extreme values, resulting in a significant drop in efficiency (the third row of Figure \ref{fig:1d_combined}). These outlying values arise from the challenges of predicting residuals when dealing with limited data. In such cases, even a slight change in the training data can impact the model's predictive power substantially, leading to poor residual predictions and, consequently, potentially inefficient PIs. Since these outliers occurred under various noise types and data sizes, it is difficult to predict exactly when they will arise or if they will occur at all.  Although improving the underlying model might be able to help mitigate the problem, such improvements are not guaranteed. It also may not be feasible to intensively train a second model for residual prediction. Therefore, if resources allow, it is advisable to repeat the experiments. Based on our experiments, the probability of encountering outliers twice consecutively is relatively low. However, the most important consideration is to remain aware that this issue can arise when using normNCMs.

There are some limitations to this work that suggest directions for future research. Firstly, 
GP has model-specific NCMs that could potentially improve the CP performance \cite{Papadopoulos2023-bj}. However, it remains unclear whether this model-specific NCM is effective for small datasets under different noise types and is a topic worth addressing in a future investigation. Secondly, the underlying models used in our experiments were not intensively tuned. While this did not impact our primary analysis, which focused on comparing the performance of different NCMs under varying conditions, it would be interesting to assess how these results compare to those obtained with extensively tuned models. Notably, the improvement in efficiency slowed down at a data size of around 400 for multi-dimensional data, suggesting that further efficiency gains might require model tuning rather than simply increasing data size or selecting different NCMs. This raises questions regarding how much additional data and the proper choice of NCMs can improve efficiency. Lastly, using smaller calibration sets introduced more variability in predictions, consistent with the findings of \cite{Linusson2014-db}. This variability is particularly evident with small datasets, highlighting the trade-off between calibration set size and prediction stability. A significant limitation in practical scenarios is the reduced amount of data available for training, especially when the dataset is already small. Given that this aspect was not a primary focus of our work, it warrants further investigation in future studies to fully understand its impact and potential solutions.

% \end{enumerate}
\section{Conclusion}
This work highlights the importance of selecting an appropriate NCM, which becomes increasingly crucial under data-scarce scenarios.  Our experiments indicate that no single NCM consistently performs best across all scenarios. The specific nature of the data significantly influences the NCM's effectiveness. As a result, proposing a clear guideline for NCM selection is challenging, given the variability in data characteristics across different contexts. Therefore, understanding how different NCMs can be adapted to particular data characteristics is essential. It is this kind of knowledge, that can positively contribute to the success of CP in real-world applications, both when dealing with abundant and scarce data.

\appendix
\section{Trade-off analysis for multi-dimensional datasets}
\label{appendix: all results}

The following figures illustrate the trade-off between validity and normalized efficiency for two-dimensional and eight-dimensional datasets under different noise types, NCMs, and target coverage rates. 

\begin{figure*}[htbp]
  \centering
  \includegraphics[width=\linewidth]{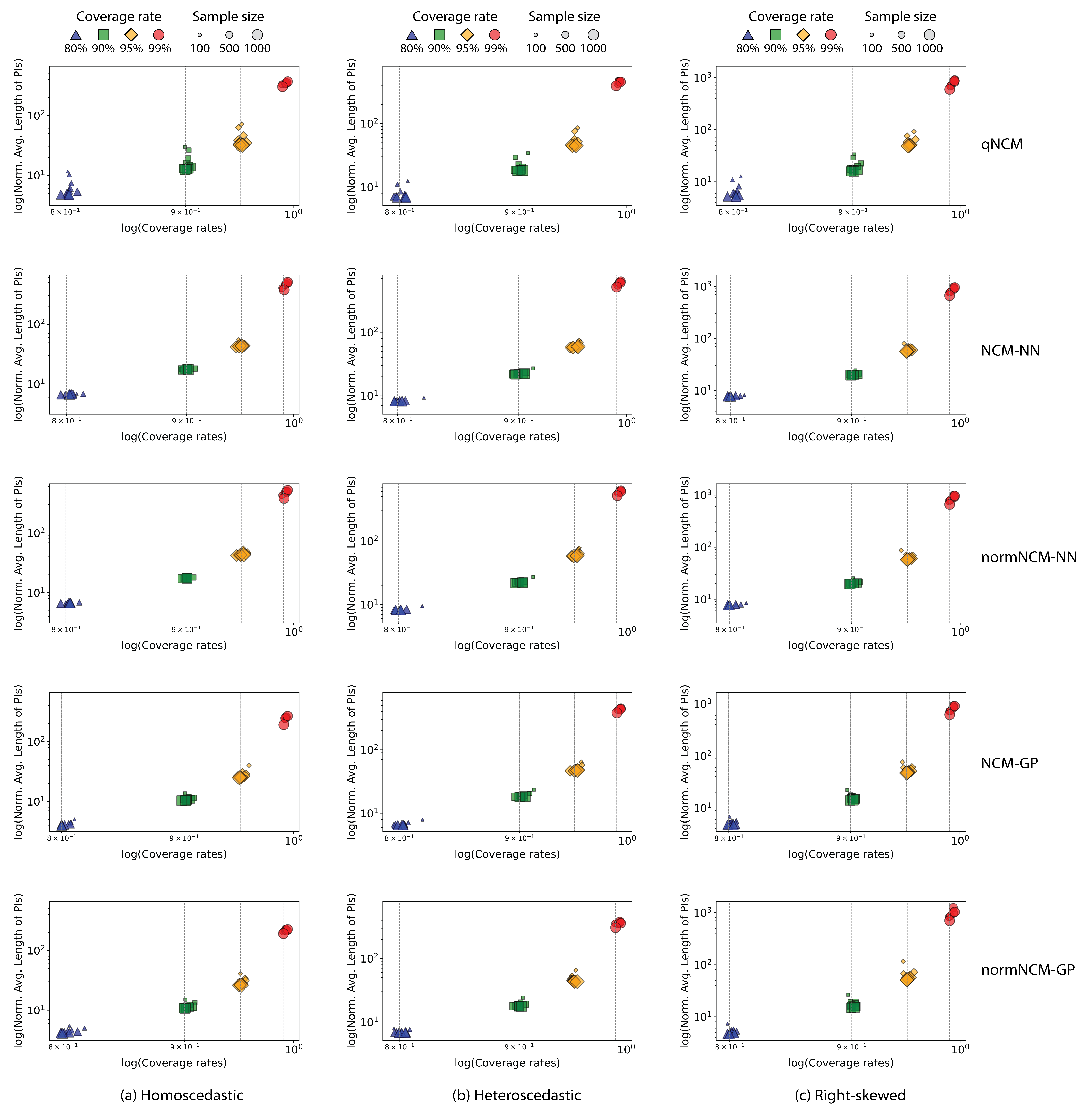}
  \caption{Trade-off between validity and normalized efficiency. In this figure, the x-axis represents validity, while the y-axis shows efficiency normalized by validity, both plotted on a logarithmic scale. The data correspond to the two-dimensional case using homoscedastic Gaussian noise (right column), heteroscedastic Gaussian noise (middle column), and right-skewed noise (left column), across varying target coverage rates and data sizes. Similar to Figure \ref{fig:1d_combined}, different shapes and sizes represent different target coverage rates and sample sizes. Each row corresponds to a different NCM: starting from the top, the rows correspond to qNCM, NCM-NN, normNCM-NN, NCM-GP, and normNCM-GP.}
\label{fig:2d-combined
}
  \label{fig:2d-combined}
\end{figure*}

\begin{figure*}[htbp]
  \centering
  \includegraphics[width=\linewidth]{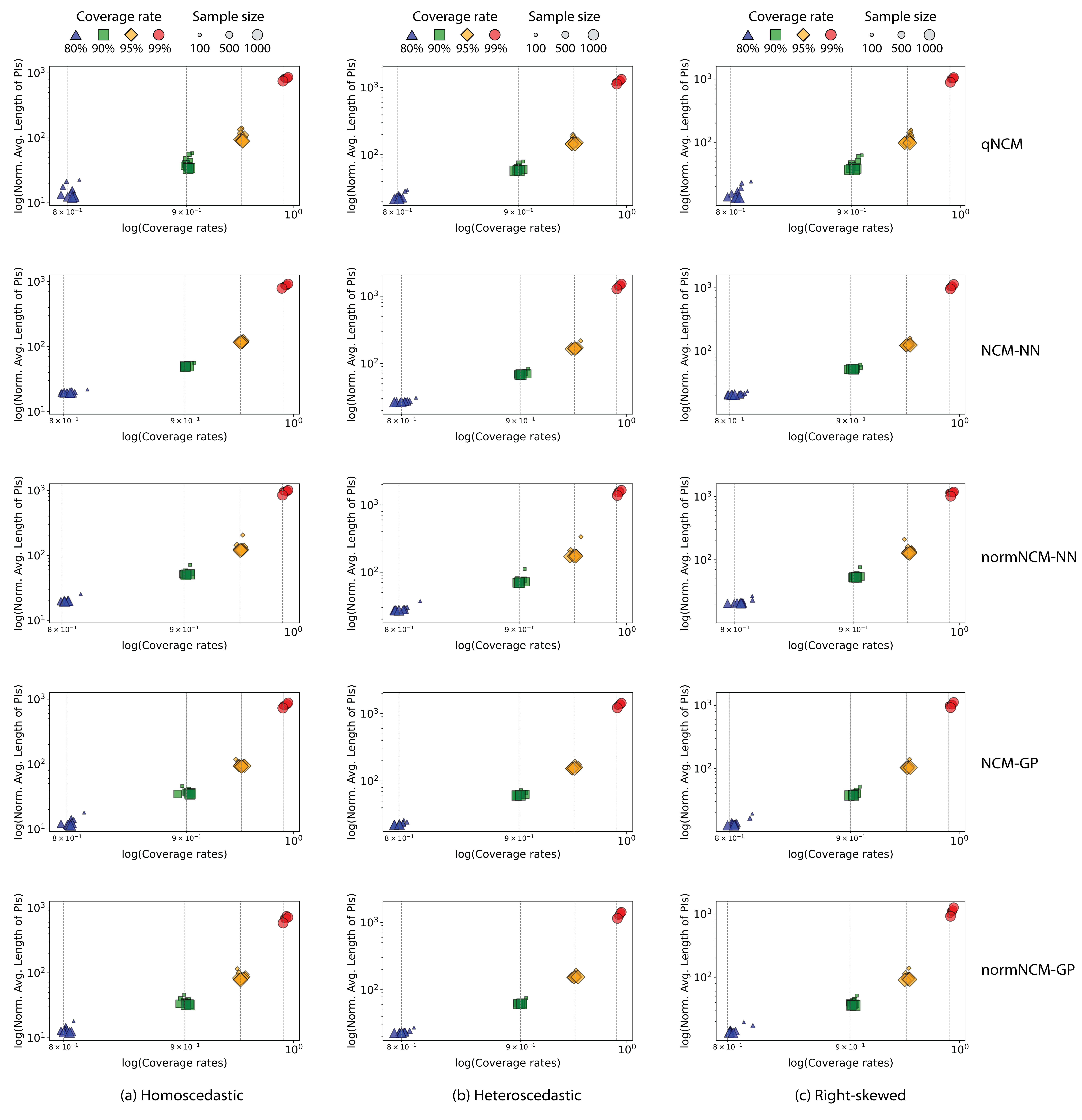}
  \caption{Trade-off between validity and normalized efficiency. In this figure, the x-axis represents validity, while the y-axis shows efficiency normalized by validity, both plotted on a logarithmic scale. The data correspond to the eight-dimensional case using homoscedastic Gaussian noise (right column), heteroscedastic Gaussian noise (middle column), and right-skewed noise (left column), across varying target coverage rates and data sizes. Similar to Figure \ref{fig:1d_combined}, different shapes and sizes represent different target coverage rates and sample sizes. Each row corresponds to a different NCM: starting from the top, the rows correspond to qNCM, NCM-NN, normNCM-NN, NCM-GP, and normNCM-GP.
}
  \label{fig:8d-combined}
\end{figure*}






% \end{thebibliography}
\bibliographystyle{elsarticle-num}
\bibliography{paperpile}
\end{document}